\documentclass{elsarticle}

\usepackage{hyperref,tikz,amsmath}
\usepackage{geometry,amssymb}
\DeclareMathOperator*{\argmax}{\text{arg\,max}}

\def\xVelocity{v_x}\def\xxa{$\xVelocity$}
\def\yVelocity{v_y}\def\xxb{$\yVelocity$}       
\def\xAcceleration{a_x}\def\xxc{$\xAcceleration$}
\def\yAcceleration{a_y}\def\xxd{$\yAcceleration$}
  
\def\distPred{\Delta x^\text{P}}\def\xxpa{$\distPred$}       
\def\diffSpeedPred{\Delta v_x^\text{P}}\def\xxpb{$\diffSpeedPred$}
\def\timeGapPred{T^\text{P}}\def\xxpc{$\timeGapPred$}
\def\accPred{a^\text{P}_x}\def\xxpd{$\accPred$}
\def\classPred{C^\text{P}}\def\xxpe{$\classPred$} 
\def\distFollo{\Delta x^\text{F}}\def\xxfa{$\distFollo$}   
\def\diffSpeedFollo{\Delta v_x^\text{F}}\def\xxfb{$\diffSpeedFollo$}   
\def\timeGapFollo{T^\text{F}}\def\xxfc{$\timeGapFollo$}     
\def\accFollo{a^\text{F}_x}\def\xxfd{$\accFollo$}         
\def\classFollo{C^\text{F}}\def\xxfe{$\classFollo$}        
\def\distPredAdj{\Delta x^\text{PA}}\def\xxpaa{$\distPredAdj$}    
\def\diffSpeedPredAdj{\Delta v_x^\text{PA}}\def\xxpab{$\diffSpeedPredAdj$} 
\def\timeGapPredAdj{T^\text{PA}}\def\xxpac{$\timeGapPredAdj$}    
\def\accPredAdj{a^\text{PA}_x}\def\xxpad{$\accPredAdj$}
\def\classPredAdj{C^\text{PA}}\def\xxpae{$\classPredAdj$}
\def\distFolloAdj{\Delta x^\text{FA}}\def\xxfaa{$\distFolloAdj$}     
\def\diffSpeedFolloAdj{\Delta v_x^\text{FA}}\def\xxfab{$\diffSpeedFolloAdj$}
\def\timeGapFolloAdj{T^\text{FA}}\def\xxfac{$\timeGapFolloAdj$}
\def\accFolloAdj{a^\text{FA}_x}\def\xxfad{$\accFolloAdj$}
\def\classFolloAdj{C^\text{FA}}\def\xxfae{$\classFolloAdj$}

\journal{Physica A}

\begin{document}

\begin{frontmatter}

\title{{\bf Predicting highway lane-changing maneuvers: A benchmark analysis of machine and ensemble learning algorithms}}

\author{Basma Khelfa\fnref{bk}}
%\address{School of Mechanical and Safety Engineering, University of Wuppertal, Germany}
\fntext[bk]{khelfa@uni-wuppertal.de}

\author{Ibrahima Ba\fnref{ib}}
%\address{School of Mechanical and Safety Engineering, University of Wuppertal, Germany}
\fntext[ib]{ibrahima.ba@uni-wuppertal.de}

\author{Antoine Tordeux\fnref{at}}
\address{School of Mechanical Engineering and Safety Engineering, University of Wuppertal, Wuppertal, Germany}
\fntext[at]{tordeux@uni-wuppertal.de}

\begin{abstract}
Understanding and predicting highway lane-change maneuvers is essential for driving modeling and its automation. The development of data-based lane-changing decision-making algorithms is nowadays in full expansion. We compare empirically in this article different machine and ensemble learning classification techniques to the MOBIL rule-based model using trajectory data of European two-lane highways. The analysis relies on instantaneous measurements of up to twenty-four spatial-temporal variables with the four neighboring vehicles on current and adjacent lanes. Preliminary descriptive investigations by principal component and logistic analyses allow identifying main variables intending a driver to change lanes. We predict two types of discretionary lane-change maneuvers: overtaking (from the slow to the fast lane) and fold-down (from the fast to the slow lane). The prediction accuracy is quantified using total, lane-changing and lane-keeping errors and associated receiver operating characteristic curves. The benchmark analysis includes logistic model, linear discriminant, decision tree, na\"ive Bayes classifier, support vector machine, neural network machine learning algorithms, and up to ten bagging and stacking ensemble learning meta-heuristics. If the rule-based model provides limited predicting accuracy, the data-based algorithms, devoid of modeling bias, allow significant prediction improvements. Cross validations show that selected neural networks and stacking algorithms allow predicting from a single observation both fold-down and overtaking maneuvers up to four seconds in advance with high accuracy.
\bigskip
\end{abstract}

\end{frontmatter}

\newpage

\section*{Highlights}
\begin{itemize}
    \item Overtaking and fold-down lane-changing maneuver predictions on European highways by bootstrap and cross-validation methods
    \item Benchmark analysis of up to six data-based algorithms, ten bagging and stacking ensemble learning meta-heuristics combining the different algorithms, and the MOBIL rule-based model
    \item Predictions of both fold-down and overtaking maneuvers up to four seconds in advance with a total accuracy larger than 98\% for selected algorithms
\end{itemize}

\section{Introduction}

Nowadays, understanding lane-change intention on highways is a matter of great interest to many researchers. 
Indeed, lane changes are one of the main causes of road accidents on highways \cite{wang1994lane, rodemerk2012development} and seem to have a major impact on traffic breakdown and oscillation phenomena \cite{cassidy2005increasing, jin2013multi, laval2008microscopic}. 
Predicting lane-changing maneuvers is also relevant to driving automation. 
Yet, the decision to change lanes on highways is a complex dynamical process in interaction with several surrounding vehicles. 
Its prediction is challenging \cite{Zhou2008}.
In lane-changing situations, drivers recognize and anticipate the driving behaviors of the surrounding vehicles by temporarily adapting their behavior, e.g., by reducing the time gap during lane-changing maneuvers or by facilitating the insertion of a new vehicle.  
Current adaptive cruise control (ACC) driver-assistance systems can behave in unrealistic ways in case of lane change due to tracking disruption, causing unsafe braking or inadequate spacing \cite{Zhou2008,CARVALHO2015,Zhang18}. 
At level L3 of driving automation on highways without supervision, coupling adaptive cruise control systems to lane-changing intent detection is crucial to anticipate the surrounding vehicle movements and regulate the speed in consequence \cite{Zhou2008}. 
Nowadays, multi-lane cooperative ACC systems forecasting lane-changing events  %and regulating the speed in safe and comfortable ways 
are in full expansion
\cite{shladover2015cooperative,schmied2016comfort,schmied2016scenario,liu2018modeling,zheng2019cooperative,mahdinia2020safety}. 
Many studies rely on vehicle-to-vehicle (V2V) communications sharing lane-changing intents in real-time \cite{shladover2012impacts,stanger2013model,shladover2015cooperative,schmidt2017cooperative}. 
V2V communications enable limited anticipation possibilities and may even be subject to communication latency. 
Recent insights in sensor technology and deep learning techniques make nowadays possible autonomous predictions of lane-changing intents several seconds in advance  \cite{kumar2013learning,dang2017time,izquierdo2017vehicle,shou2020long,mozaffari2022early}. 

Data-based approaches rely on classification algorithms and large amounts of data. 
Many studies use NGSIM trajectory data-sets \cite{ngsim} collected from 2007 to 2010 on US highways or data-sets collected on Chinese highways (see, e.g., Ubiquitous Traffic Eyes data-set \cite{ute} or \cite{zheng2014recent,guo2018empirical}).
Lane changes in Chinese and US traffic are symmetric. 
However, two different lane-changing maneuvers must be distinguished in European highway traffic imposing to overtake by the left: overtaking (from right to left lanes) and fold-down (from left to right lanes). 
Few studies consider the asymmetry between overtaking and fold-down maneuvers.
In this article, we aim to carry out a benchmark analysis of different models and data-based algorithms for the prediction of lane change on European highways using HighD trajectory data-set \cite{highd}. 
The predictions rely on instantaneous measurements of up to twenty-four spatial-temporal variables with the four neighboring vehicles on current and adjacent lanes.
The comparison includes rule-based and statistical models, machine learning algorithms, and ensemble learning meta-heuristics. 
Seventeen approaches are quantitatively analyzed using total, lane-keeping, lane-changing errors, and associated ROC curves.  
The analysis is done separately for overtaking and fold-down maneuvers.
Cross-validated forecasting are tested up to five seconds before a vehicle effectively changes lanes.
The results mainly show that different mechanisms operate in cases of overtaking (from the slow to the fast lane) and fold-down (from the fast to the slow lane) lane changes. 
Furthermore, data-based algorithms outperform the predictions of the rule-based algorithm.
Selected neural networks and stacking ensemble learning algorithms allow
predicting both fold-down and overtaking maneuvers up to four seconds in advance with a total accuracy larger than 98\%.

The structure of the article is the following. 
In section~\ref{section2}, we present a chronological bibliographical review including rule-based models and data-based algorithms for the prediction of highway lane-changing maneuvers. 
The collection of lane-changing maneuvers and associated measurements are reported in section~\ref{section3}, as well as a preliminary descriptive analysis based on principal component and logistic regression.
The quantitative benchmark analysis of prediction results obtained with the MOBIL rule-based model and different machine and ensemble learning algorithms is detailed in section~\ref{section4}. 
Section~\ref{section5} shows a summary of the results and some concluding remarks.

\section{Bibliographical review}
\label{section2}

In the literature, several distinct methods allow predicting the lane change intention. 
Chronologically, first attempts rely on rule-based approaches such as the kinematic lane-changing model by Gipps \cite{gipps1986} in the middle of the 1980's or, more recently, the \textit{minimising overall braking induced by lane change} (MOBIL) model by Kesting et al.\ \cite{kesting2007}. 
Empirical approaches based on large amounts of data instead of a theory appeared more recently. 
Early data-driven approaches rely on linear statistical classification algorithms such as logistic regression, linear discriminant analysis, or na\"ive Bayes classifier. 
Currently, non-linear machine, deep, or ensemble learning techniques, e.g., random forest, support vector machine, or neural network algorithms, are under active development. 
Ensemble learning approaches such as bagging, stacking, and boosting methods combining different algorithms are also beginning to be implemented. 
They seem to provide the most accurate results. 
We briefly review the rule-based and data-based approaches for lane-change prediction by focusing on recent machine and ensemble learning approaches.
See \cite{moridpour2010lane,dogan2011autonomous,rahman2013review, choi2021Comparison,das2021Machine} for detailed dedicated surveys on the topic.

\subsection{Rule-based models} 

Rule-based models aim to identify fundamental underlying mechanisms initiating a driver to change lanes.
The kinematic model by Gipps \cite{gipps1986} takes into consideration a combination of five criteria: keeping safe gap distances, controlling vehicle speed, maintaining steering motion intended by the driver, identifying positions of static obstacles, and checking for the existence of transit lanes or heavy vehicles.  
The lane-changing model by Ahmed et al.\ \cite{ahmed96} is a similar physics-based approach considering both mandatory and discretionary lane change events through critical lead and lag gaps. 
Such approaches are currently called \textit{gap acceptance} models in the literature. 
This model class is generalized and calibrated in the light of real data in \cite{toledo2003,toledo2007}. 
In \cite{hidas2005,hidas2002} a multi-agent traffic simulation platform uses a cognitive approach based on the time gap and a driver reaction time to model lane change on freeways. 
Similar approaches rely on game theory lane, gains induced by changing lanes, and possibilities of collisions \cite{kita99}. 

The gap acceptance model by Gipps is one of the first approaches for the modeling of lane-changing behavior. 
However, the most widely used approach to microscopic multi-lane traffic modeling is the rule-based model MOBIL \cite{kesting2007}. 
The model relies on accelerations on current and adjacent lanes determined using a car-following model based on distance and speed difference. 
Coupled with factors of aggressiveness and politeness, the decision to change lanes lies in acceleration difference gains by changing lanes. 
Similar formulation based on cost functions have been developed thereafter (see, e.g., \cite{wei2010prediction}). 
The forms of the rule-based and physics-based lane-changing models differ from one approach to another.
However, distance spacing and speed difference with the neighboring vehicles on current and adjacent lanes are systematically input variables.

\subsection{Data-based algorithms}

In contrast to gap acceptance and rule-based models, data-based algorithms predict lane-change maneuvers with no modelling assumption nor identification of fundamental underlying mechanisms. 
The predictions rely on large amounts of data used to train different types of classification algorithms. 
However, similarly to rule-based models, inputs are in most cases distance, speed difference, and other time-space variables with the surrounding vehicles on current and adjacent lanes.
Prominent examples of classification algorithms used to predict lane changes on highways are logistic regression, linear discriminant, na\"ive Bayes classifier, support vector machine, neural network, among other statistical models and machine learning techniques.
See, e.g., the dedicated reviews on the topic \cite{dogan2011autonomous,choi2021Comparison,das2021Machine,deshmukh2022overview}. 

A large variety of statistical methods, nowadays generically related as machine learning techniques, allows automatic pattern recognition in high dimensional settings \cite{aeberhard1994comparative}. 
Such algorithms are currently used to classify highway driving situations and predict lane-changing maneuvers.
For instance, linear discriminant analyses allow forecasting lane-keeping or lane-changing decision-making in \cite{yang2018driving}.
Support vector machines enable from trajectories of current neighboring vehicles to predict lane changes in \cite{mandalia2005,hanw2017}. 
Such approaches rely on linear algorithms.
Some authors also use non-linear decision trees for the prediction of lane-changing maneuvers \cite{hou2013modeling,motamedidehkordi2017modeling}. 
In \cite{dou2016}, a method combining support vector machine and artificial neural network predicts discretionary and mandatory lane-changing from the auxiliary to the adjacent lanes. 
The input parameters are distances and speed differences with the surrounding vehicles. 
Many authors used hidden Markov models and na\"ive Bayes classifiers to predict the lane changes few seconds before they occur \cite{hou2013modeling,carvalho2016,yuan2018lane,xu2019hybrid,jin2020gauss,khelfa2021lane}.
Generally speaking, non-linear algorithms by artificial neural networks \cite{tomar2010prediction,zheng2014predicting,bae2020cooperation} and especially convolutional neural networks \cite{mozaffari2022early,xu2021recognition}, deep learning including feedback mechanisms (e.g., long short term memory (LSTM) algorithm) \cite{dang2017time,zou2019predicting,shou2020long,wirthmuller2021predicting,li2021lane,long2022does,shi2022integrated}, or encoder-decoder models \cite{ren2022method,wei2022fine} provide the most accurate predictions.
More recent approaches rely on ensemble learning meta-heuristics combining several algorithms for the prediction.
For instance, in \cite{bejani2018context} an ensemble learning based on decision tree, support vector machine, multi-layer perceptron, and K-nearest neighbors determine the maneuvers. 
Other approaches rely on gradient boosting heuristics \cite{xing2020ensemble,zhang2022xgboost}.
Note that ensemble learning techniques are used in other domains of traffic engineering, e.g., for injury severity prediction in traffic accidents \cite{cuenca2018traffic}.

Many studies comparatively analyze different machine learning techniques for the prediction of lane-changing maneuvers on highways.  
For instance, Dou et al.\ \cite{dou2016} compare support vector machines and artificial neural networks. 
Authors consider Bayes classifier and decision trees as well \cite{hou2013modeling}, or Gaussian classification, support vector classification, and neural network \cite{rakos2020lane}. 
Recent reviews and surveys cover more algorithms again (see, e.g., \cite{choi2021Comparison,das2021Machine,deshmukh2022overview}). 
In this article, we carry out an empirical benchmark analysis including up to six machine learning algorithms and ten ensemble learning meta-heuristics. 
The analysis also includes MOBIL as a reference for rule-based models.

\section{Data collection and preliminary analysis}
\label{section3}
In the following, we present the HighD highway trajectory data-set \cite{highd} and the process of collecting lane-changing maneuvers and variables of interest. 
We also present preliminary descriptive visualizations of the data relying on correlation and principal component analysis. 
A logistic analysis allows identifying some of the main variables influencing lane-changing decision-making.

\subsection{Collection of maneuvers and explanatory variables}

\begin{figure}[!ht]
\centering
\input{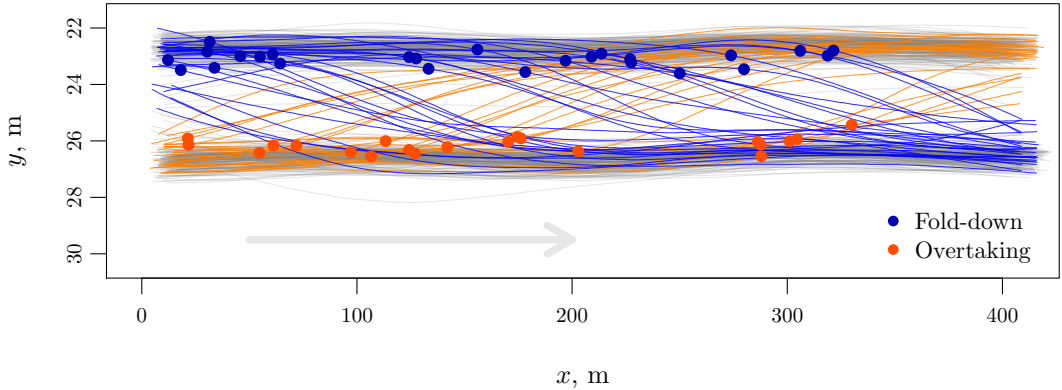}\vspace{-5mm}
\caption{Example of trajectories on a two-lane highway (record \texttt{02\_Tracks}). The points indicate lane-change decision-making two seconds before the vehicle crosses the highway center-line. For readability, $x$ and $y$ axes do not have the same scale.}
\label{figTraj}\medskip
\end{figure}

The data used to compare different algorithms predicting lane-changing maneuvers on highways comes from the HighD data-set \cite{highd}.
The HighD data-set is a recent, freely available collection of naturalistic vehicle trajectories. 
The trajectories were recorded using a drone in 2017 and 2018 on 420 meters portions of German highways with up to 5 lanes. 
The data-set includes 110\,500 trajectories (cars and trucks), over 44\,500 kilometers of bi-directional two-lane and three-lane highways, and 147 driving hours. 
The traffic was recorded at six different locations with different traffic conditions. 
We select 13 records corresponding to two-lane highways and extract from the trajectories the lane-keeping and the lane-changing maneuvers of the cars. 
Different instantaneous space-time variables are measured for each observed car.
The measurement for lane-changing maneuvers occurs two seconds before the vehicle crosses the highway center-line. 
Figure~\ref{figTraj} draws an example of trajectories on two-lane highways (sample record \texttt{02\_Tracks}). % with punctual lane-changing decision-making.

\begin{figure}[!ht]
\centering\medskip
\input{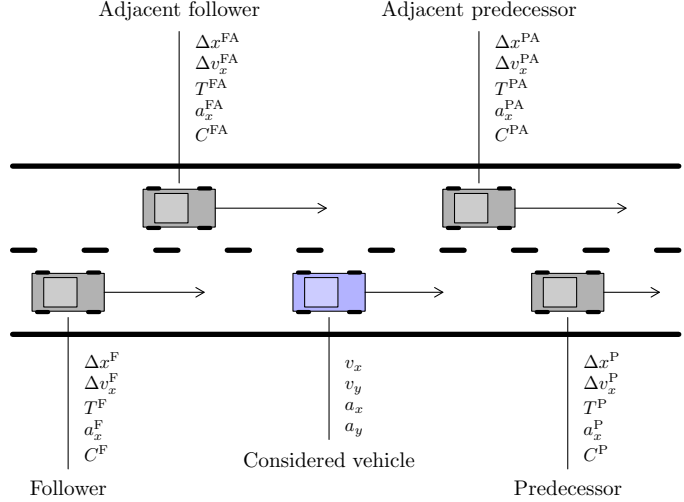}\vspace{-2mm}
\caption{Four surrounding vehicles on a two-lane highway and measured variables (see Table~\protect\ref{tablePredicators} for details).}
\label{figPredicators}
\end{figure}

We distinguish in the following four types of maneuvers on European two-lane highways:
\begin{enumerate}
    \item Lane-keeping on the right lane (LKR);
    \item Lane-keeping on the left lane (LKL);
    \item Lane-changing from the left to the right lane (fold-down maneuver FD);
    \item Lane-changing from the right to the left lane (overtaking maneuver OV).
\end{enumerate}
To understand and predict car lane-keeping and lane-changing decision-making, we measure up to 24 space-time variables with the four surrounding vehicles in the current and adjacent lanes (see Fig.~\ref{figPredicators} and Table~\ref{tablePredicators}). 
The variables that most influence the lane-changing intent are the speed of current vehicles and the speed difference and spacing with the four surrounding vehicles on current and adjacent lanes. 
These variables are the inputs of the MOBIL rule-based model \cite{kesting2007} and most of the data-based approaches as well.

\begin{table}[!ht]\centering\small\bigskip
\renewcommand{\arraystretch}{1.4}\def\sl{\\[1mm]}
\begin{tabular}{l|lc}
Variables&Description&Unit\\
\hline
\multicolumn{3}{l}{Considered vehicle}\\
\hline\\[-3.5mm]
$M\in\{1,2,3,4\}$&Maneuver (LKR, LKL, FD, OV)&---\sl
$\xVelocity=\frac{x(t+\delta t/2)-x(t-\delta t/2)}{\delta t}$&Longitudinal velocity&m/s\sl
$\yVelocity=\frac{y(t+\delta t/2)-y(t-\delta t/2)}{\delta t}$&Lateral velocity&m/s\sl
$\xAcceleration=\frac{\xVelocity(t+\delta t/2)-\xVelocity(t-\delta t/2)}{\delta t}$&Longitudinal acceleration&m/s$^2$\sl
$\yAcceleration=\frac{\yVelocity(t+\delta t/2)-\yVelocity(t-\delta t/2)}{\delta t}$&Lateral acceleration&m/s$^2$\\[1mm]
%$\class$&Vehicle class (car or truck)$\qquad\qquad\qquad\qquad$&---\\
\multicolumn{3}{l}{For all 4 surrounding vehicles $\text{S}=\{\text{P},\text{F},\text{PA},\text{FA}\}$, see Fig.~\protect\ref{figPredicators}}\\
\hline\\[-3.5mm]
$\Delta x^\text{S}=|x^\text{S}-x|$&Spacing&m\sl
$\Delta v_x^\text{S}=\xVelocity^\text{S}-\xVelocity$&Speed difference&m/s\sl
$T^\text{S}=\frac{\Delta x^\text{S}-(\ell+\ell^\text{S})/2}{\xVelocity}$&Time gap&s\sl
$a^\text{S}_x=\frac{\xVelocity^\text{S}(t+\delta t/2)-\xVelocity^\text{S}(t-\delta t/2)}{\delta t}$&Longitudinal acceleration&m/s$^2$\sl
$C^\text{S}\in\{0,1\}$&Vehicle class (car or truck)&---
\end{tabular}\smallskip
\caption{Descriptions and units of the twenty-five measured variables. The maneuver $M$ is the variable to explain, while the twenty-four others measurements are the explanatory variables. Here $\delta t=0.1$\;s while $\ell$ is the vehicle length in meters. Note that the spacing variables are systematically positive.}
\label{tablePredicators}
\end{table}

\subsection{Preliminary data analysis}

\begin{figure}[!ht]
\centering\smallskip
\input{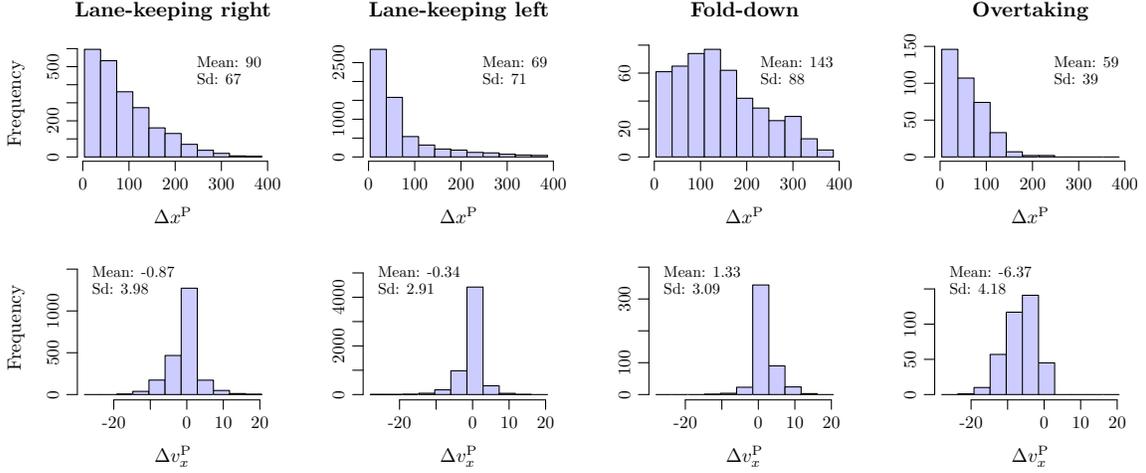}\vspace{-7mm}
\caption{Histograms of spacing (in meter) and speed difference (in meter per second) with the predecessor for the four possible maneuvers.}
\label{figHist}
\end{figure}

We first preliminary analyze the data using descriptive statistics.
It is worth noting that 24\% and 67\% of the observed cars keep the right (slow) and left (fast) lanes, respectively, see Table~\ref{tableStat}. 
lane-changing maneuvers only represent 9\% of the observed car maneuvers.
Indeed, the left lane is mainly used by trucks (94\% of the observations) with low speed (the maximal allowed speed for trucks being 25\,ms$^{-1}$ while there is no maximal legal speed for the cars).  
Fig.~\ref{figHist} presents the histograms of the spacing and speed difference with the predecessor according to the four maneuvers. 
Different patterns can be identified. 
The spacing distribution is compact when overtaking (mean spacing of 59\,m). 
The distribution is also quite concentrated when keeping the left lane compared to the distribution on the right lane (mean spacing of 69 and 90\,m, respectively). 
However, surprisingly, the distribution is the most spread out in case of fold-down (mean spacing of 143\,m). 
Indeed, drivers tend to fold down when the traffic situation is free.
The speed differences reveal more polarised patterns again. 
The distributions are centered when keeping the lane (mean speed difference of -0.87 and -0.34\,ms$^{-1}$ on left and right lanes, respectively). 
The speed differences tend to be positive when folding down (mean of 1.33\,ms$^{-1}$) and negative when overtaking (mean of -6.37\,ms$^{-1}$).
These patterns suggest that highway lane-changing maneuvers can be predicted from distances and speed differences with surrounding vehicles in the current and adjacent lanes.

\begin{table}[!ht]\medskip\centering\small
\renewcommand{\arraystretch}{1.5}
\begin{tabular}{r|cccccccc}
&$\overline{\xVelocity}$ (ms$^{-1}$)&$\overline{\distPred}$ (m)&$\overline{\timeGapPred}$ (s)~~~&\#\,LKR&\#\,LKL&\#\,FD&\#\,OV&\footnotesize$\sum$\\
\hline
Car&32.17&77.85&2.36~~~&2191&6074&489&371&9125\\[-1mm]
Truck&24.16&92.26&3.80~~~&2132&80&28&23&2263\\[-1mm]
\multicolumn{4}{r}{\footnotesize$\sum$}&4323&6154&517&394&11388\\[-1mm]
\multicolumn{4}{r}{\footnotesize$\%$}&37.96&54.04&4.54&3.46&100.00
\end{tabular}\vspace{-2mm}
\caption{Mean speed, mean distance with the predecessor, mean time gap, and frequency statistics for lane-keeping and lane-changing maneuvers of cars and trucks on two-lane highways.}
\label{tableStat}
\end{table}

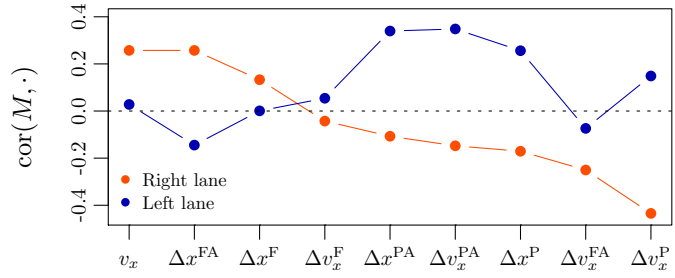
\begin{figure}[!ht]
\centering
% Created by tikzDevice version 0.12.3 on 2020-12-03 23:36:38
% !TEX encoding = UTF-8 Unicode
\begin{tikzpicture}[x=.9pt,y=.9pt]
\definecolor{fillColor}{RGB}{255,255,255}
\path[use as bounding box,fill=fillColor,fill opacity=0.00] (0,0) rectangle (308.85,138.98);
\begin{scope}
\path[clip] ( 48.00, 24.00) rectangle (284.85,114.98);
\definecolor{drawColor}{RGB}{255,77,0}

\path[draw=drawColor,line width= 0.4pt,line join=round,line cap=round] ( 62.77, 97.47) -- ( 78.18, 97.47);

\path[draw=drawColor,line width= 0.4pt,line join=round,line cap=round] ( 89.66, 95.01) -- (106.12, 87.61);

\path[draw=drawColor,line width= 0.4pt,line join=round,line cap=round] (116.66, 81.93) -- (133.95, 70.96);

\path[draw=drawColor,line width= 0.4pt,line join=round,line cap=round] (144.85, 66.38) -- (160.58, 62.72);

\path[draw=drawColor,line width= 0.4pt,line join=round,line cap=round] (172.36, 60.49) -- (187.90, 58.21);

\path[draw=drawColor,line width= 0.4pt,line join=round,line cap=round] (199.82, 56.84) -- (215.27, 55.55);

\path[draw=drawColor,line width= 0.4pt,line join=round,line cap=round] (227.01, 53.39) -- (242.90, 48.82);

\path[draw=drawColor,line width= 0.4pt,line join=round,line cap=round] (253.66, 43.83) -- (271.08, 32.23);
\definecolor{fillColor}{RGB}{255,77,0}

\path[fill=fillColor] ( 56.77, 97.48) circle (  2.25);

\path[fill=fillColor] ( 84.18, 97.46) circle (  2.25);

\path[fill=fillColor] (111.60, 85.15) circle (  2.25);

\path[fill=fillColor] (139.01, 67.74) circle (  2.25);

\path[fill=fillColor] (166.42, 61.36) circle (  2.25);

\path[fill=fillColor] (193.84, 57.34) circle (  2.25);

\path[fill=fillColor] (221.25, 55.05) circle (  2.25);

\path[fill=fillColor] (248.66, 47.16) circle (  2.25);

\path[fill=fillColor] (276.07, 28.91) circle (  2.25);
\end{scope}
\begin{scope}
\path[clip] (  0.00,  0.00) rectangle (308.85,138.98);
\definecolor{drawColor}{RGB}{0,0,0}

\path[draw=drawColor,line width= 0.4pt,line join=round,line cap=round] ( 48.00, 32.33) -- ( 48.00,111.61);

\path[draw=drawColor,line width= 0.4pt,line join=round,line cap=round] ( 48.00, 32.33) -- ( 42.00, 32.33);

\path[draw=drawColor,line width= 0.4pt,line join=round,line cap=round] ( 48.00, 52.15) -- ( 42.00, 52.15);

\path[draw=drawColor,line width= 0.4pt,line join=round,line cap=round] ( 48.00, 71.97) -- ( 42.00, 71.97);

\path[draw=drawColor,line width= 0.4pt,line join=round,line cap=round] ( 48.00, 91.79) -- ( 42.00, 91.79);

\path[draw=drawColor,line width= 0.4pt,line join=round,line cap=round] ( 48.00,111.61) -- ( 42.00,111.61);

\node[text=drawColor,rotate= 90.00,anchor=base,inner sep=0pt, outer sep=0pt, scale=  0.80] at ( 38.40, 32.33) {-0.4};

\node[text=drawColor,rotate= 90.00,anchor=base,inner sep=0pt, outer sep=0pt, scale=  0.80] at ( 38.40, 52.15) {-0.2};

\node[text=drawColor,rotate= 90.00,anchor=base,inner sep=0pt, outer sep=0pt, scale=  0.80] at ( 38.40, 71.97) {0.0};

\node[text=drawColor,rotate= 90.00,anchor=base,inner sep=0pt, outer sep=0pt, scale=  0.80] at ( 38.40, 91.79) {0.2};

\node[text=drawColor,rotate= 90.00,anchor=base,inner sep=0pt, outer sep=0pt, scale=  0.80] at ( 38.40,111.61) {0.4};

\path[draw=drawColor,line width= 0.4pt,line join=round,line cap=round] ( 48.00, 24.00) --
	(284.85, 24.00) --
	(284.85,114.98) --
	( 48.00,114.98) --
	( 48.00, 24.00);
\end{scope}
\begin{scope}
\path[clip] (  0.00,  0.00) rectangle (308.85,138.98);
\definecolor{drawColor}{RGB}{0,0,0}

\node[text=drawColor,rotate= 90.00,anchor=base,inner sep=0pt, outer sep=0pt, scale=  1.00] at ( 15.60, 69.49) {$\mbox{cor}(M,\cdot)$};
\end{scope}
\begin{scope}
\path[clip] (  0.00,  0.00) rectangle (308.85,138.98);
\definecolor{drawColor}{RGB}{0,0,0}

\path[draw=drawColor,line width= 0.4pt,line join=round,line cap=round] ( 56.77, 24.00) -- ( 56.77, 24.00);

\path[draw=drawColor,line width= 0.4pt,line join=round,line cap=round] ( 56.77, 24.00) -- ( 56.77, 18.00);

\node[text=drawColor,anchor=base,inner sep=0pt, outer sep=0pt, scale=  0.80] at ( 56.77,  7.20) {\xxa};

\path[draw=drawColor,line width= 0.4pt,line join=round,line cap=round] ( 84.18, 24.00) -- ( 84.18, 24.00);

\path[draw=drawColor,line width= 0.4pt,line join=round,line cap=round] ( 84.18, 24.00) -- ( 84.18, 18.00);

\node[text=drawColor,anchor=base,inner sep=0pt, outer sep=0pt, scale=  0.80] at ( 84.18,  7.20) {\xxfaa};

\path[draw=drawColor,line width= 0.4pt,line join=round,line cap=round] (111.60, 24.00) -- (111.60, 24.00);

\path[draw=drawColor,line width= 0.4pt,line join=round,line cap=round] (111.60, 24.00) -- (111.60, 18.00);

\node[text=drawColor,anchor=base,inner sep=0pt, outer sep=0pt, scale=  0.80] at (111.60,  7.20) {\xxfa};

\path[draw=drawColor,line width= 0.4pt,line join=round,line cap=round] (139.01, 24.00) -- (139.01, 24.00);

\path[draw=drawColor,line width= 0.4pt,line join=round,line cap=round] (139.01, 24.00) -- (139.01, 18.00);

\node[text=drawColor,anchor=base,inner sep=0pt, outer sep=0pt, scale=  0.80] at (139.01,  7.20) {\xxfb};

\path[draw=drawColor,line width= 0.4pt,line join=round,line cap=round] (166.42, 24.00) -- (166.42, 24.00);

\path[draw=drawColor,line width= 0.4pt,line join=round,line cap=round] (166.42, 24.00) -- (166.42, 18.00);

\node[text=drawColor,anchor=base,inner sep=0pt, outer sep=0pt, scale=  0.80] at (166.42,  7.20) {\xxpaa};

\path[draw=drawColor,line width= 0.4pt,line join=round,line cap=round] (193.84, 24.00) -- (193.84, 24.00);

\path[draw=drawColor,line width= 0.4pt,line join=round,line cap=round] (193.84, 24.00) -- (193.84, 18.00);

\node[text=drawColor,anchor=base,inner sep=0pt, outer sep=0pt, scale=  0.80] at (193.84,  7.20) {\xxpab};

\path[draw=drawColor,line width= 0.4pt,line join=round,line cap=round] (221.25, 24.00) -- (221.25, 24.00);

\path[draw=drawColor,line width= 0.4pt,line join=round,line cap=round] (221.25, 24.00) -- (221.25, 18.00);

\node[text=drawColor,anchor=base,inner sep=0pt, outer sep=0pt, scale=  0.80] at (221.25,  7.20) {\xxpa};

\path[draw=drawColor,line width= 0.4pt,line join=round,line cap=round] (248.66, 24.00) -- (248.66, 24.00);

\path[draw=drawColor,line width= 0.4pt,line join=round,line cap=round] (248.66, 24.00) -- (248.66, 18.00);

\node[text=drawColor,anchor=base,inner sep=0pt, outer sep=0pt, scale=  0.80] at (248.66,  7.20) {\xxfab};

\path[draw=drawColor,line width= 0.4pt,line join=round,line cap=round] (276.07, 24.00) -- (276.07, 24.00);

\path[draw=drawColor,line width= 0.4pt,line join=round,line cap=round] (276.07, 24.00) -- (276.07, 18.00);

\node[text=drawColor,anchor=base,inner sep=0pt, outer sep=0pt, scale=  0.80] at (276.07,  7.20) {\xxpb};
\end{scope}
\begin{scope}
\path[clip] ( 48.00, 24.00) rectangle (284.85,114.98);
\definecolor{drawColor}{RGB}{0,0,179}

\path[draw=drawColor,line width= 0.4pt,line join=round,line cap=round] ( 61.86, 71.57) -- ( 79.09, 60.82);

\path[draw=drawColor,line width= 0.4pt,line join=round,line cap=round] ( 89.50, 60.43) -- (106.29, 69.25);

\path[draw=drawColor,line width= 0.4pt,line join=round,line cap=round] (117.49, 73.18) -- (133.12, 76.20);

\path[draw=drawColor,line width= 0.4pt,line join=round,line cap=round] (143.19, 81.65) -- (162.25,101.32);

\path[draw=drawColor,line width= 0.4pt,line join=round,line cap=round] (172.42,105.81) -- (187.84,106.29);

\path[draw=drawColor,line width= 0.4pt,line join=round,line cap=round] (199.53,104.58) -- (215.56, 99.24);

\path[draw=drawColor,line width= 0.4pt,line join=round,line cap=round] (225.10, 92.75) -- (244.81, 69.25);

\path[draw=drawColor,line width= 0.4pt,line join=round,line cap=round] (253.34, 68.42) -- (271.40, 82.95);
\definecolor{fillColor}{RGB}{0,0,179}

\path[fill=fillColor] ( 56.77, 74.74) circle (  2.25);

\path[fill=fillColor] ( 84.18, 57.64) circle (  2.25);

\path[fill=fillColor] (111.60, 72.04) circle (  2.25);

\path[fill=fillColor] (139.01, 77.34) circle (  2.25);

\path[fill=fillColor] (166.42,105.62) circle (  2.25);

\path[fill=fillColor] (193.84,106.48) circle (  2.25);

\path[fill=fillColor] (221.25, 97.34) circle (  2.25);

\path[fill=fillColor] (248.66, 64.66) circle (  2.25);

\path[fill=fillColor] (276.07, 86.71) circle (  2.25);
\definecolor{drawColor}{RGB}{0,0,0}

\path[draw=drawColor,line width= 0.4pt,dash pattern=on 1pt off 3pt ,line join=round,line cap=round] ( 48.00, 71.97) -- (284.85, 71.97);
\definecolor{fillColor}{RGB}{255,77,0}

\path[fill=fillColor] ( 55.20, 43.20) circle (  1.80);
\definecolor{fillColor}{RGB}{0,0,179}

\path[fill=fillColor] ( 55.20, 33.60) circle (  1.80);

\node[text=drawColor,anchor=base west,inner sep=0pt, outer sep=0pt, scale=  0.70] at ( 62.40, 40.45) {Right lane};

\node[text=drawColor,anchor=base west,inner sep=0pt, outer sep=0pt, scale=  0.70] at ( 62.40, 30.85) {Left lane};
\end{scope}
\end{tikzpicture}\vspace{-2mm}
\caption{Correlation coefficients between the maneuver, the speed, and the distance and speed difference with the four surrounding vehicles in the right and left lanes.}
\label{figCor}\medskip
\end{figure}

\begin{figure}[!ht]
\centering\vspace{-8mm}
\input{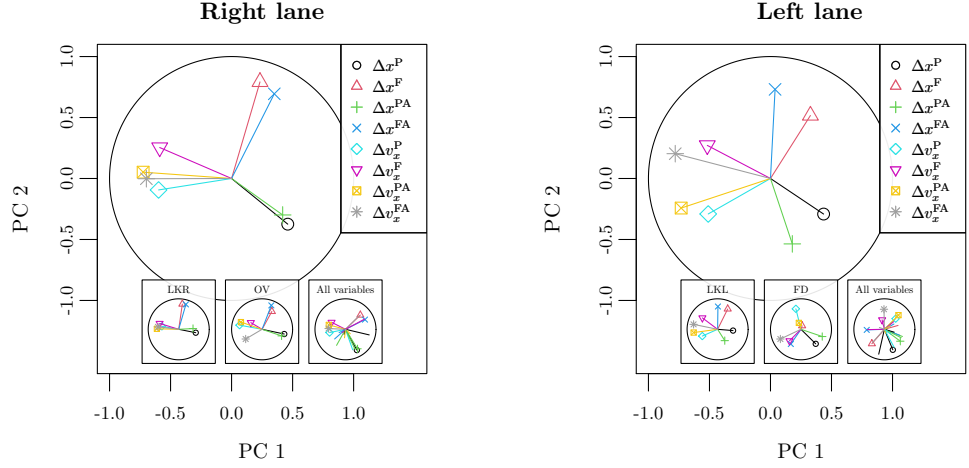}\vspace{-4mm}
\caption{Correlation circles for the two first principal components of the measured variables in the right lane (overtaking maneuver, left panel) and in the left lane (fold-down maneuver, right panel).}
\label{figPCA}
\end{figure}

Figure~\ref{figCor} presents the correlation coefficients between the maneuvers (lane-keeping and lane-changing) in the right and left lanes and the speed of the considered vehicle, speed difference and spacing with the four surrounding vehicles. 
Overtaking occurs for weak speed difference and spacing with the predecessor and high speed and distance with the follower on the adjacent lane. 
In contrast, fold-down maneuver mainly relies on high spacing and speed difference with the predecessor on the adjacent lane. 
Generally speaking, the correlations for overtaking and fold-down are opposed.  
This is especially the case for the predecessor spacing and speed difference. 
The results suggest that different mechanisms operate according to the maneuver. 
The correlation circles over the two first principal components drawn in Fig.~\ref{figPCA} confirm such a statement.
Overtaking maneuvers result from relatively simple mechanisms with only three parameters, while fold-down maneuvers are more complicated behaviors, including more variables for decision-making. 
Note that the two first components contain 45\% of the variability with eight variables and 27\% with all the twenty-four variables.

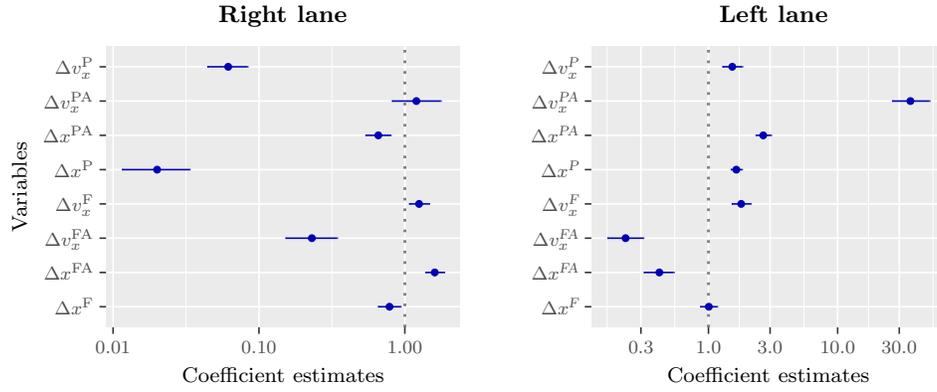
\begin{figure}[!ht]
\centering\bigskip
{\bf\small\hspace{10mm} Right lane \hspace{47mm} Left lane}\\[.5mm]
{\footnotesize% Created by tikzDevice version 0.12.3 on 2021-01-19 22:34:14
% !TEX encoding = UTF-8 Unicode
\begin{tikzpicture}[x=.9pt,y=.9pt]
\definecolor{fillColor}{RGB}{255,255,255}
\path[use as bounding box,fill=fillColor,fill opacity=0.00] (0,0) rectangle (200.75,154.42);
\begin{scope}
\path[clip] (  0.00,  0.00) rectangle (200.75,154.42);
\definecolor{drawColor}{RGB}{255,255,255}
\definecolor{fillColor}{RGB}{255,255,255}

\path[draw=drawColor,line width= 0.6pt,line join=round,line cap=round,fill=fillColor] (  0.00,  0.00) rectangle (200.75,154.42);
\end{scope}
\begin{scope}
\path[clip] ( 46.38, 30.69) rectangle (195.25,148.92);
\definecolor{fillColor}{gray}{0.92}

\path[fill=fillColor] ( 46.38, 30.69) rectangle (195.25,148.92);
\definecolor{drawColor}{RGB}{255,255,255}

\path[draw=drawColor,line width= 0.3pt,line join=round] ( 79.99, 30.69) --
	( 79.99,148.92);

\path[draw=drawColor,line width= 0.3pt,line join=round] (141.31, 30.69) --
	(141.31,148.92);

\path[draw=drawColor,line width= 0.6pt,line join=round] ( 46.38, 39.34) --
	(195.25, 39.34);

\path[draw=drawColor,line width= 0.6pt,line join=round] ( 46.38, 53.76) --
	(195.25, 53.76);

\path[draw=drawColor,line width= 0.6pt,line join=round] ( 46.38, 68.18) --
	(195.25, 68.18);

\path[draw=drawColor,line width= 0.6pt,line join=round] ( 46.38, 82.59) --
	(195.25, 82.59);

\path[draw=drawColor,line width= 0.6pt,line join=round] ( 46.38, 97.01) --
	(195.25, 97.01);

\path[draw=drawColor,line width= 0.6pt,line join=round] ( 46.38,111.43) --
	(195.25,111.43);

\path[draw=drawColor,line width= 0.6pt,line join=round] ( 46.38,125.85) --
	(195.25,125.85);

\path[draw=drawColor,line width= 0.6pt,line join=round] ( 46.38,140.27) --
	(195.25,140.27);

\path[draw=drawColor,line width= 0.6pt,line join=round] ( 49.33, 30.69) --
	( 49.33,148.92);

\path[draw=drawColor,line width= 0.6pt,line join=round] (110.65, 30.69) --
	(110.65,148.92);

\path[draw=drawColor,line width= 0.6pt,line join=round] (171.96, 30.69) --
	(171.96,148.92);
\definecolor{drawColor}{gray}{0.50}

\path[draw=drawColor,line width= 1.1pt,dash pattern=on 1pt off 3pt ,line join=round] (171.96, 30.69) -- (171.96,148.92);
\definecolor{drawColor}{RGB}{0,0,179}

\path[draw=drawColor,line width= 0.6pt,line join=round] ( 81.48, 97.01) --
	( 81.48, 97.01);

\path[draw=drawColor,line width= 0.6pt,line join=round] ( 81.48, 97.01) --
	( 53.14, 97.01);

\path[draw=drawColor,line width= 0.6pt,line join=round] ( 53.14, 97.01) --
	( 53.14, 97.01);

\path[draw=drawColor,line width= 0.6pt,line join=round] (170.16, 39.34) --
	(170.16, 39.34);

\path[draw=drawColor,line width= 0.6pt,line join=round] (170.16, 39.34) --
	(160.72, 39.34);

\path[draw=drawColor,line width= 0.6pt,line join=round] (160.72, 39.34) --
	(160.72, 39.34);

\path[draw=drawColor,line width= 0.6pt,line join=round] (165.94,111.43) --
	(165.94,111.43);

\path[draw=drawColor,line width= 0.6pt,line join=round] (165.94,111.43) --
	(155.41,111.43);

\path[draw=drawColor,line width= 0.6pt,line join=round] (155.41,111.43) --
	(155.41,111.43);

\path[draw=drawColor,line width= 0.6pt,line join=round] (188.48, 53.76) --
	(188.48, 53.76);

\path[draw=drawColor,line width= 0.6pt,line join=round] (188.48, 53.76) --
	(180.63, 53.76);

\path[draw=drawColor,line width= 0.6pt,line join=round] (180.63, 53.76) --
	(180.63, 53.76);

\path[draw=drawColor,line width= 0.6pt,line join=round] (105.84,140.27) --
	(105.84,140.27);

\path[draw=drawColor,line width= 0.6pt,line join=round] (105.84,140.27) --
	( 88.95,140.27);

\path[draw=drawColor,line width= 0.6pt,line join=round] ( 88.95,140.27) --
	( 88.95,140.27);

\path[draw=drawColor,line width= 0.6pt,line join=round] (182.19, 82.59) --
	(182.19, 82.59);

\path[draw=drawColor,line width= 0.6pt,line join=round] (182.19, 82.59) --
	(173.75, 82.59);

\path[draw=drawColor,line width= 0.6pt,line join=round] (173.75, 82.59) --
	(173.75, 82.59);

\path[draw=drawColor,line width= 0.6pt,line join=round] (187.02,125.85) --
	(187.02,125.85);

\path[draw=drawColor,line width= 0.6pt,line join=round] (187.02,125.85) --
	(166.58,125.85);

\path[draw=drawColor,line width= 0.6pt,line join=round] (166.58,125.85) --
	(166.58,125.85);

\path[draw=drawColor,line width= 0.6pt,line join=round] (143.46, 68.18) --
	(143.46, 68.18);

\path[draw=drawColor,line width= 0.6pt,line join=round] (143.46, 68.18) --
	(121.85, 68.18);

\path[draw=drawColor,line width= 0.6pt,line join=round] (121.85, 68.18) --
	(121.85, 68.18);
\definecolor{fillColor}{RGB}{0,0,179}

\path[draw=drawColor,line width= 0.4pt,line join=round,line cap=round,fill=fillColor] ( 67.84, 97.01) circle (  1.43);

\path[draw=drawColor,line width= 0.4pt,line join=round,line cap=round,fill=fillColor] (165.51, 39.34) circle (  1.43);

\path[draw=drawColor,line width= 0.4pt,line join=round,line cap=round,fill=fillColor] (160.81,111.43) circle (  1.43);

\path[draw=drawColor,line width= 0.4pt,line join=round,line cap=round,fill=fillColor] (184.52, 53.76) circle (  1.43);

\path[draw=drawColor,line width= 0.4pt,line join=round,line cap=round,fill=fillColor] ( 97.72,140.27) circle (  1.43);

\path[draw=drawColor,line width= 0.4pt,line join=round,line cap=round,fill=fillColor] (177.94, 82.59) circle (  1.43);

\path[draw=drawColor,line width= 0.4pt,line join=round,line cap=round,fill=fillColor] (176.80,125.85) circle (  1.43);

\path[draw=drawColor,line width= 0.4pt,line join=round,line cap=round,fill=fillColor] (132.88, 68.18) circle (  1.43);
\end{scope}
\begin{scope}
\path[clip] (  0.00,  0.00) rectangle (200.75,154.42);
\definecolor{drawColor}{gray}{0.30}

\node[text=drawColor,anchor=base east,inner sep=0pt, outer sep=0pt, scale=  0.88] at ( 41.43, 36.31) {\xxfa};

\node[text=drawColor,anchor=base east,inner sep=0pt, outer sep=0pt, scale=  0.88] at ( 41.43, 50.73) {\xxfaa};

\node[text=drawColor,anchor=base east,inner sep=0pt, outer sep=0pt, scale=  0.88] at ( 41.43, 65.15) {\xxfab};

\node[text=drawColor,anchor=base east,inner sep=0pt, outer sep=0pt, scale=  0.88] at ( 41.43, 79.56) {\xxfb};

\node[text=drawColor,anchor=base east,inner sep=0pt, outer sep=0pt, scale=  0.88] at ( 41.43, 93.98) {\xxpa};

\node[text=drawColor,anchor=base east,inner sep=0pt, outer sep=0pt, scale=  0.88] at ( 41.43,108.40) {\xxpaa};

\node[text=drawColor,anchor=base east,inner sep=0pt, outer sep=0pt, scale=  0.88] at ( 41.43,122.82) {\xxpab};

\node[text=drawColor,anchor=base east,inner sep=0pt, outer sep=0pt, scale=  0.88] at ( 41.43,137.24) {\xxpb};
\end{scope}
\begin{scope}
\path[clip] (  0.00,  0.00) rectangle (200.75,154.42);
\definecolor{drawColor}{gray}{0.20}

\path[draw=drawColor,line width= 0.6pt,line join=round] ( 43.63, 39.34) --
	( 46.38, 39.34);

\path[draw=drawColor,line width= 0.6pt,line join=round] ( 43.63, 53.76) --
	( 46.38, 53.76);

\path[draw=drawColor,line width= 0.6pt,line join=round] ( 43.63, 68.18) --
	( 46.38, 68.18);

\path[draw=drawColor,line width= 0.6pt,line join=round] ( 43.63, 82.59) --
	( 46.38, 82.59);

\path[draw=drawColor,line width= 0.6pt,line join=round] ( 43.63, 97.01) --
	( 46.38, 97.01);

\path[draw=drawColor,line width= 0.6pt,line join=round] ( 43.63,111.43) --
	( 46.38,111.43);

\path[draw=drawColor,line width= 0.6pt,line join=round] ( 43.63,125.85) --
	( 46.38,125.85);

\path[draw=drawColor,line width= 0.6pt,line join=round] ( 43.63,140.27) --
	( 46.38,140.27);
\end{scope}
\begin{scope}
\path[clip] (  0.00,  0.00) rectangle (200.75,154.42);
\definecolor{drawColor}{gray}{0.20}

\path[draw=drawColor,line width= 0.6pt,line join=round] ( 49.33, 27.94) --
	( 49.33, 30.69);

\path[draw=drawColor,line width= 0.6pt,line join=round] (110.65, 27.94) --
	(110.65, 30.69);

\path[draw=drawColor,line width= 0.6pt,line join=round] (171.96, 27.94) --
	(171.96, 30.69);
\end{scope}
\begin{scope}
\path[clip] (  0.00,  0.00) rectangle (200.75,154.42);
\definecolor{drawColor}{gray}{0.30}

\node[text=drawColor,anchor=base,inner sep=0pt, outer sep=0pt, scale=  0.88] at ( 49.33, 19.68) {0.01};

\node[text=drawColor,anchor=base,inner sep=0pt, outer sep=0pt, scale=  0.88] at (110.65, 19.68) {0.10};

\node[text=drawColor,anchor=base,inner sep=0pt, outer sep=0pt, scale=  0.88] at (171.96, 19.68) {1.00};
\end{scope}
\begin{scope}
\path[clip] (  0.00,  0.00) rectangle (200.75,154.42);
\definecolor{drawColor}{RGB}{0,0,0}

\node[text=drawColor,anchor=base,inner sep=0pt, outer sep=0pt, scale=  1.0] at (120.81,  7.64) {Coefficient estimates};
\end{scope}
\begin{scope}
\path[clip] (  0.00,  0.00) rectangle (200.75,154.42);
\definecolor{drawColor}{RGB}{0,0,0}

\node[text=drawColor,rotate= 90.00,anchor=base,inner sep=0pt, outer sep=0pt, scale=  1.0] at ( 13.08, 89.80) {Variables};
\end{scope}
\end{tikzpicture}\sl% Created by tikzDevice version 0.12.3 on 2021-01-19 22:34:21
% !TEX encoding = UTF-8 Unicode
\begin{tikzpicture}[x=.9pt,y=.9pt]
\definecolor{fillColor}{RGB}{255,255,255}
\path[use as bounding box,fill=fillColor,fill opacity=0.00] (0,0) rectangle (200.75,154.42);
\begin{scope}
\path[clip] (  0.00,  0.00) rectangle (200.75,154.42);
\definecolor{drawColor}{RGB}{255,255,255}
\definecolor{fillColor}{RGB}{255,255,255}

\path[draw=drawColor,line width= 0.6pt,line join=round,line cap=round,fill=fillColor] (  0.00,  0.00) rectangle (200.75,154.42);
\end{scope}
\begin{scope}
\path[clip] ( 46.38, 30.69) rectangle (195.25,148.92);
\definecolor{fillColor}{gray}{0.92}

\path[fill=fillColor] ( 46.38, 30.69) rectangle (195.25,148.92);
\definecolor{drawColor}{RGB}{255,255,255}

\path[draw=drawColor,line width= 0.3pt,line join=round] ( 53.02, 30.69) --
	( 53.02,148.92);

\path[draw=drawColor,line width= 0.3pt,line join=round] ( 81.41, 30.69) --
	( 81.41,148.92);

\path[draw=drawColor,line width= 0.3pt,line join=round] (108.55, 30.69) --
	(108.55,148.92);

\path[draw=drawColor,line width= 0.3pt,line join=round] (135.70, 30.69) --
	(135.70,148.92);

\path[draw=drawColor,line width= 0.3pt,line join=round] (162.84, 30.69) --
	(162.84,148.92);

\path[draw=drawColor,line width= 0.3pt,line join=round] (189.99, 30.69) --
	(189.99,148.92);

\path[draw=drawColor,line width= 0.6pt,line join=round] ( 46.38, 39.34) --
	(195.25, 39.34);

\path[draw=drawColor,line width= 0.6pt,line join=round] ( 46.38, 53.76) --
	(195.25, 53.76);

\path[draw=drawColor,line width= 0.6pt,line join=round] ( 46.38, 68.18) --
	(195.25, 68.18);

\path[draw=drawColor,line width= 0.6pt,line join=round] ( 46.38, 82.59) --
	(195.25, 82.59);

\path[draw=drawColor,line width= 0.6pt,line join=round] ( 46.38, 97.01) --
	(195.25, 97.01);

\path[draw=drawColor,line width= 0.6pt,line join=round] ( 46.38,111.43) --
	(195.25,111.43);

\path[draw=drawColor,line width= 0.6pt,line join=round] ( 46.38,125.85) --
	(195.25,125.85);

\path[draw=drawColor,line width= 0.6pt,line join=round] ( 46.38,140.27) --
	(195.25,140.27);

\path[draw=drawColor,line width= 0.6pt,line join=round] ( 67.22, 30.69) --
	( 67.22,148.92);

\path[draw=drawColor,line width= 0.6pt,line join=round] ( 95.60, 30.69) --
	( 95.60,148.92);

\path[draw=drawColor,line width= 0.6pt,line join=round] (121.50, 30.69) --
	(121.50,148.92);

\path[draw=drawColor,line width= 0.6pt,line join=round] (149.89, 30.69) --
	(149.89,148.92);

\path[draw=drawColor,line width= 0.6pt,line join=round] (175.79, 30.69) --
	(175.79,148.92);
\definecolor{drawColor}{gray}{0.50}

\path[draw=drawColor,line width= 1.1pt,dash pattern=on 1pt off 3pt ,line join=round] ( 95.60, 30.69) -- ( 95.60,148.92);
\definecolor{drawColor}{RGB}{0,0,179}

\path[draw=drawColor,line width= 0.6pt,line join=round] (109.67, 97.01) --
	(109.67, 97.01);

\path[draw=drawColor,line width= 0.6pt,line join=round] (109.67, 97.01) --
	(104.95, 97.01);

\path[draw=drawColor,line width= 0.6pt,line join=round] (104.95, 97.01) --
	(104.95, 97.01);

\path[draw=drawColor,line width= 0.6pt,line join=round] ( 99.23, 39.34) --
	( 99.23, 39.34);

\path[draw=drawColor,line width= 0.6pt,line join=round] ( 99.23, 39.34) --
	( 92.17, 39.34);

\path[draw=drawColor,line width= 0.6pt,line join=round] ( 92.17, 39.34) --
	( 92.17, 39.34);

\path[draw=drawColor,line width= 0.6pt,line join=round] (121.89,111.43) --
	(121.89,111.43);

\path[draw=drawColor,line width= 0.6pt,line join=round] (121.89,111.43) --
	(115.44,111.43);

\path[draw=drawColor,line width= 0.6pt,line join=round] (115.44,111.43) --
	(115.44,111.43);

\path[draw=drawColor,line width= 0.6pt,line join=round] ( 80.99, 53.76) --
	( 80.99, 53.76);

\path[draw=drawColor,line width= 0.6pt,line join=round] ( 80.99, 53.76) --
	( 68.44, 53.76);

\path[draw=drawColor,line width= 0.6pt,line join=round] ( 68.44, 53.76) --
	( 68.44, 53.76);

\path[draw=drawColor,line width= 0.6pt,line join=round] (109.79,140.27) --
	(109.79,140.27);

\path[draw=drawColor,line width= 0.6pt,line join=round] (109.79,140.27) --
	(101.48,140.27);

\path[draw=drawColor,line width= 0.6pt,line join=round] (101.48,140.27) --
	(101.48,140.27);

\path[draw=drawColor,line width= 0.6pt,line join=round] (113.31, 82.59) --
	(113.31, 82.59);

\path[draw=drawColor,line width= 0.6pt,line join=round] (113.31, 82.59) --
	(105.38, 82.59);

\path[draw=drawColor,line width= 0.6pt,line join=round] (105.38, 82.59) --
	(105.38, 82.59);

\path[draw=drawColor,line width= 0.6pt,line join=round] (188.48,125.85) --
	(188.48,125.85);

\path[draw=drawColor,line width= 0.6pt,line join=round] (188.48,125.85) --
	(172.84,125.85);

\path[draw=drawColor,line width= 0.6pt,line join=round] (172.84,125.85) --
	(172.84,125.85);

\path[draw=drawColor,line width= 0.6pt,line join=round] ( 68.24, 68.18) --
	( 68.24, 68.18);

\path[draw=drawColor,line width= 0.6pt,line join=round] ( 68.24, 68.18) --
	( 53.14, 68.18);

\path[draw=drawColor,line width= 0.6pt,line join=round] ( 53.14, 68.18) --
	( 53.14, 68.18);
\definecolor{fillColor}{RGB}{0,0,179}

\path[draw=drawColor,line width= 0.4pt,line join=round,line cap=round,fill=fillColor] (107.31, 97.01) circle (  1.43);

\path[draw=drawColor,line width= 0.4pt,line join=round,line cap=round,fill=fillColor] ( 95.75, 39.34) circle (  1.43);

\path[draw=drawColor,line width= 0.4pt,line join=round,line cap=round,fill=fillColor] (118.63,111.43) circle (  1.43);

\path[draw=drawColor,line width= 0.4pt,line join=round,line cap=round,fill=fillColor] ( 74.96, 53.76) circle (  1.43);

\path[draw=drawColor,line width= 0.4pt,line join=round,line cap=round,fill=fillColor] (105.61,140.27) circle (  1.43);

\path[draw=drawColor,line width= 0.4pt,line join=round,line cap=round,fill=fillColor] (109.35, 82.59) circle (  1.43);

\path[draw=drawColor,line width= 0.4pt,line join=round,line cap=round,fill=fillColor] (180.48,125.85) circle (  1.43);

\path[draw=drawColor,line width= 0.4pt,line join=round,line cap=round,fill=fillColor] ( 60.77, 68.18) circle (  1.43);
\end{scope}
\begin{scope}
\path[clip] (  0.00,  0.00) rectangle (200.75,154.42);
\definecolor{drawColor}{gray}{0.30}

\node[text=drawColor,anchor=base east,inner sep=0pt, outer sep=0pt, scale=  0.88] at ( 41.43, 36.31) {\xxfa};

\node[text=drawColor,anchor=base east,inner sep=0pt, outer sep=0pt, scale=  0.88] at ( 41.43, 50.73) {\xxfaa};

\node[text=drawColor,anchor=base east,inner sep=0pt, outer sep=0pt, scale=  0.88] at ( 41.43, 65.15) {\xxfab};

\node[text=drawColor,anchor=base east,inner sep=0pt, outer sep=0pt, scale=  0.88] at ( 41.43, 79.56) {\xxfb};

\node[text=drawColor,anchor=base east,inner sep=0pt, outer sep=0pt, scale=  0.88] at ( 41.43, 93.98) {\xxpa};

\node[text=drawColor,anchor=base east,inner sep=0pt, outer sep=0pt, scale=  0.88] at ( 41.43,108.40) {\xxpaa};

\node[text=drawColor,anchor=base east,inner sep=0pt, outer sep=0pt, scale=  0.88] at ( 41.43,122.82) {\xxpab};

\node[text=drawColor,anchor=base east,inner sep=0pt, outer sep=0pt, scale=  0.88] at ( 41.43,137.24) {\xxpb};
\end{scope}
\begin{scope}
\path[clip] (  0.00,  0.00) rectangle (200.75,154.42);
\definecolor{drawColor}{gray}{0.20}

\path[draw=drawColor,line width= 0.6pt,line join=round] ( 43.63, 39.34) --
	( 46.38, 39.34);

\path[draw=drawColor,line width= 0.6pt,line join=round] ( 43.63, 53.76) --
	( 46.38, 53.76);

\path[draw=drawColor,line width= 0.6pt,line join=round] ( 43.63, 68.18) --
	( 46.38, 68.18);

\path[draw=drawColor,line width= 0.6pt,line join=round] ( 43.63, 82.59) --
	( 46.38, 82.59);

\path[draw=drawColor,line width= 0.6pt,line join=round] ( 43.63, 97.01) --
	( 46.38, 97.01);

\path[draw=drawColor,line width= 0.6pt,line join=round] ( 43.63,111.43) --
	( 46.38,111.43);

\path[draw=drawColor,line width= 0.6pt,line join=round] ( 43.63,125.85) --
	( 46.38,125.85);

\path[draw=drawColor,line width= 0.6pt,line join=round] ( 43.63,140.27) --
	( 46.38,140.27);
\end{scope}
\begin{scope}
\path[clip] (  0.00,  0.00) rectangle (200.75,154.42);
\definecolor{drawColor}{gray}{0.20}

\path[draw=drawColor,line width= 0.6pt,line join=round] ( 67.22, 27.94) --
	( 67.22, 30.69);

\path[draw=drawColor,line width= 0.6pt,line join=round] ( 95.60, 27.94) --
	( 95.60, 30.69);

\path[draw=drawColor,line width= 0.6pt,line join=round] (121.50, 27.94) --
	(121.50, 30.69);

\path[draw=drawColor,line width= 0.6pt,line join=round] (149.89, 27.94) --
	(149.89, 30.69);

\path[draw=drawColor,line width= 0.6pt,line join=round] (175.79, 27.94) --
	(175.79, 30.69);
\end{scope}
\begin{scope}
\path[clip] (  0.00,  0.00) rectangle (200.75,154.42);
\definecolor{drawColor}{gray}{0.30}

\node[text=drawColor,anchor=base,inner sep=0pt, outer sep=0pt, scale=  0.88] at ( 67.22, 19.68) {\normalfont 0.3};

\node[text=drawColor,anchor=base,inner sep=0pt, outer sep=0pt, scale=  0.88] at ( 95.60, 19.68) {\normalfont 1.0};

\node[text=drawColor,anchor=base,inner sep=0pt, outer sep=0pt, scale=  0.88] at (121.50, 19.68) {\normalfont 3.0};

\node[text=drawColor,anchor=base,inner sep=0pt, outer sep=0pt, scale=  0.88] at (149.89, 19.68) {\normalfont 10.0};

\node[text=drawColor,anchor=base,inner sep=0pt, outer sep=0pt, scale=  0.88] at (175.79, 19.68) {\normalfont 30.0};
\end{scope}
\begin{scope}
\path[clip] (  0.00,  0.00) rectangle (200.75,154.42);
\definecolor{drawColor}{RGB}{0,0,0}

\node[text=drawColor,anchor=base,inner sep=0pt, outer sep=0pt, scale=  1.0] at (120.81,  7.64) {\normalfont Coefficient estimates};
\end{scope}
\begin{scope}
\path[clip] (  0.00,  0.00) rectangle (200.75,154.42);
\definecolor{drawColor}{RGB}{0,0,0}

\node[text=drawColor,rotate= 90.00,anchor=base,inner sep=0pt, outer sep=0pt, scale=  1.0] at ( 13.08, 89.80) {~};
\end{scope}
\end{tikzpicture}}\vspace{-2mm}
\caption{Exponential estimates by maximum likelihood of the logistic coefficients for the distances and speed differences with the four surrounding vehicles on the right lane (overtaking maneuver, left panel) and on the left lane (fold-down maneuver, right panel).}
\label{figGally}\medskip
\end{figure}

Next, we perform a logistic analysis to understand the role of the speed difference and distance variables with the four surrounding vehicles in the lane-changing decision-making.  
The objective is to identify the most significant factors intending a driver to change lanes. 
Logistic regression provides insight into lane-changing maneuvers by interpreting the statistical estimates of the coefficients, see Eqs.~(\ref{logit}) and (\ref{logitb}) and \cite{khelfa2021understanding}. 
Fig.~\ref{figGally} presents the exponential map of the estimates by maximum likelihood with 95\% confidence intervals.
Exponents greater than one mean positive influences on lane-changing decision-making. 
Exponents smaller than one rely on negative influences of the variables.
While exponents close to one mean that the variables have no significant roles in the lane-changing intent.
On right lanes, decision-making for overtaking is motivated by weak spacing and speed difference with the predecessor and the speed difference with the follower on the adjacent lane.
The decision to fold down from the left lane is a more complex process influenced by combinations of all the variables except the distance to the follower. 
These observations confirm the results obtained with the correlation and principal component analyses, see Figs.~\ref{figCor} and \ref{figPCA}.

\section{Prediction results}
 \label{section4}
 
In this section, we empirically compare lane-changing predictions of the MOBIL rule-based model and different data-based algorithms.
The tested algorithms are linear and non-linear statistical models, machine learning, and ensemble learning meta-heuristics. 
We present the algorithms and the methodology before drawing the benchmark analysis.
The predictions are first performed two seconds before the vehicle crosses the central-line using longitudinal distance and speed differences between the four surrounding vehicles before providing estimates up to five seconds using all measured variables.
Note that lateral motion variables enable trivial prediction of the lane-changing maneuvers on short times. 
We first focus on the longitudinal system state for a fair comparison with the MOBIL model. 
Predictions over up to five seconds using all twenty-four measured variables, including the vehicle lateral motions, are carried out with the machine and ensemble learning meta-heuristics.

\subsection{Lane-changing model and algorithms}

The objective is to compare empirically a rule-based model for lane-changing prediction, namely the model Minimising Overall Braking Induced by Lane change (MOBIL), and different classification algorithms based on data.
The data-based algorithms are classical statistical models, e.g., logistic regression, linear discriminant, na\"ive Bayes classifier, decision tree, and machine learning algorithms, such as feed-forward artificial neural networks or support vector machine.
We compare each of the algorithms and the rule-based model.   
The model MOBIL is based on the distances and speed differences between the four neighbors in the current and adjacent lanes. 
The analysis is firstly carried out using these eight measurements as explanatory variables for the classification algorithms. 
Later in the analysis, we use all the twenty-four measured variables and ensemble learning techniques such as bootstrap aggregating (bagging) or stacking methods with different meta-heuristics to predict up to five seconds in advance lane-changing maneuvers.

\subsubsection{MOBIL model}
The microscopic model Minimising Overall Braking Induced by Lane change (MOBIL) is a rule-based reference for lane-changing prediction \cite{kesting2007}. 
The model includes two criteria: a safety criterion and an incentive criterion. 
The algorithm uses the Intelligent Driver (ID) car-following model \cite{treiber2000congested} to determine vehicle accelerations on current and intended lane according to
\begin{equation}\label{IDM}
    \left\{\begin{array}{lcl}
\displaystyle a=\alpha\left[1-\bigg(\frac{v}{v_0}\bigg)^4-\bigg(\frac{f(v_1,v)}{\Delta x-\ell}\bigg)^2\right],\\[5mm]
\displaystyle f(v,v_1)=\ell+Tv+v\frac{v-v_1}{2\sqrt{\alpha\beta}}.\end{array}\right.
\end{equation}
The ID car-following model relies on five parameters
\begin{itemize}
    \item the desired (or maximum) speed $v_0>0$,
    \item the desired time gap $T>0$,
    \item the maximum acceleration $\alpha>0$,
    \item the comfortable deceleration $\beta>0$,
    \item the vehicle length $\ell>0$.
\end{itemize}
The safety criterion authorizes the lane change beyond a minimum braking threshold. 
A lane-changing only occurs if the acceleration $\tilde a_\textsc{fa}$ of the following vehicle in the intended lane in case of lane change is higher than a minimum threshold $b_\text{safe}=-4$~m/s$^2$:
\begin{equation}\label{eq1}
\tilde{a}_\textsc{fa} \geq b_\text{safe}.
\end{equation} 

The incentive criterion measures the attractiveness of a given lane according to a performance gain and a politeness factor.
It may be symmetric (no distinction between overtaking and fold-down maneuvers) or asymmetric.
The asymmetric incentive criterion reflects the European driving rules imposing overtaking to the left.    
The criterion for overtaking is
\begin{equation}\label{eq8}
\tilde{a}- a + p_R ( \tilde{a}_\textsc{fa}- a_\textsc{fa}) > \Delta a + \Delta a_\text{bias} = b_R.
\end{equation}
For the fold-down maneuver, the criterion is
\begin{equation}\label{eq7}
\tilde{a}- a + p_L (
\tilde{a}_\textsc{f}- a_\textsc{f}) > \Delta a - \Delta a_\text{bias} = b_L.
\end{equation}
Here $a$ and $\tilde a$ are the accelerations in the current lane and by considering the lane-changing maneuver.
The intelligent driver car-following model Eq.~(\ref{IDM}) allows determining these acceleration values.
Besides the five parameters of ID model, the lane-changing rule depends on two parameters: 
\begin{itemize}
    \item the politeness proportion $p\in[0,1]$,
    \item the acceleration threshold $b$ that may be positive for overtaking or negative in case of fold-down maneuver. 
\end{itemize}
We calibrate these seven parameters by least squares over the sample of maneuver observations. 
Estimations are carried out separately on the right and left lanes. 
The optimization is done on R using the meta-heuristic \texttt{optim} based on the "L-BFGS-B" quasi-Newton iterative algorithm \cite{Byrd1995}. 
Such an algorithm allows reaching local optima.
We approximate the global optimum by sampling the optimization over 50 random initial parameter values. 
%The prediction mean square errors according to politeness proportion $p$ and the acceleration threshold $b$ parameters are shown in Fig.~\ref{figMOBIL}.
The parameter estimates are presented in Table~\ref{tableMOBIL}. % for the full (imbalanced) data set and a balanced data set containing the same lane-keeping and lane-changing observation numbers.
Estimates for the desired speed $v_0$ and vehicle length $\ell$ are physically sound and relatively stable on the right and left lanes. 
The preferred time gap $T$, acceleration $\alpha$, and deceleration $\beta$ estimates depend in turn on the lane and associated maneuver. 
The estimates are close to the ones given in \cite{treiber2000congested} for overtaking. 
While the preferred deceleration is lower (by a factor two), and the time gap and preferred acceleration are higher (by a factor four and two, respectively) in case of fold-down.
The politeness proportion $p$ and the acceleration threshold $b$ also present specific values according to the maneuvers.
The acceleration threshold $b$ is positive for overtaking and negative for fold-down maneuvers. 
The politeness proportion $p$ is slightly higher in case of fold-down.

\begin{table}[!ht]\medskip\centering\small
\renewcommand{\arraystretch}{1.5}\def\foo{\footnotesize}\def\sl{\\}
\begin{tabular}{lll|cc}
&&\multicolumn{1}{c}{}&\multicolumn{2}{c}{Estimates}\sl
&Parameter&Tested range&Right lane&Left lane\\
\hline
\foo IDM&$v_0$ \foo(m/s)&[20,80]&63.16 \foo(20.88)&58.77 \foo(22.57)\sl
&$T$ \foo(s)&[0,5]&1.04  \foo(0.53)&3.97 \foo(0.82)\sl
&$\alpha$ \foo(m/s$^2$)&[0,5]&1.45  \foo(0.93)&2.76 \foo(1.10)\sl
&$\beta$ \foo(m/s$^2$)&[0,5]&2.60  \foo(0.83)&1.37 \foo(1.08)\sl
&$\ell$ \foo(m)&[0,10]&7.27  \foo(0.93)&6.17 \foo(0.96)\\
\foo MOBIL&$p$&[0,1]&0.53  \foo(0.27)&0.64 \foo(0.27)\sl
&$b$ \foo(m/s$^2$)&[-4,4]&1.56  \foo(0.61)&-1.14 \foo(1.38)
\end{tabular}
\caption{Least square estimate for the parameters of the ID car-following and MOBIL lane-changing models. % over the full (imbalanced) sample and a balanced data set containing the same lane-keeping and lane-changing observation numbers. 
The values into brackets are the estimate standard deviations.}
\label{tableMOBIL}
\end{table}

\subsubsection{Data-based algorithms}

Many data-based methods, statistical models, and machine learning algorithms allow predicting the lane change intention. 
Prominent examples are logistic regression, decision tree, support vector machine or neural network (see, e.g., the reviews \cite{dogan2011autonomous,rahman2013review,choi2021Comparison,das2021Machine}). 
In the following, we present the algorithms used for the benchmark analysis.
Later in the analysis, we combine these different algorithms in ensemble learning meta-heuristics. 
We denote $X=(X^1,\ldots,X^p)$ as the matrix of explanatory variables and $Y$ as vector of the binary maneuver to predict (lane-changing or lane-keeping).

\paragraph{Logistic regression}
The logistic regression (LR) allows analyzing the relationships between a binary variable and one or more continuous explanatory variables. % \cite{reglog}.
The logistic model is currently used to predict the probability that an event happens. 
For an observation $i$, the model is
\begin{equation}
\pi_i=\frac{1}{1+\exp(-\alpha_0-\alpha_1X^1_i-\ldots-\alpha_p X^p_i)}
\label{logit}
\end{equation}
We estimate the coefficients $\alpha_0,\ldots,\alpha_p$ by maximising  the likelihood numerically in R (package \texttt{GLM})
\begin{equation}
L(\mathbf \alpha)=\prod_{i=1}^n\pi_i^{Y_i}(1-\pi_i)^{1-Y_i}.
\label{logitb}
\end{equation}
We assume intent of lane change if $\pi_i>0.5$ and lane-keeping for $\pi_i\le0.5$.

\paragraph{Linear discriminant analysis}
The principle of the linear discriminant analysis (LDA) method is to find a linear combination of attributes, i.e., a discriminant 
\begin{equation}
    D(\alpha,X_i)=\alpha_0+\alpha_1X^1_i+\ldots+\alpha_p X^p_i,
\end{equation}
maximizing the mean value variance between clusters (the inter-variance) or minimizing the mean variance within the clusters (the intra-variance). 
The sum of intra- and inter-variability being constant and equal to the total variability, minimizing the intra-variance is equivalent to maximizing the inter-variability. 
Under homoscedasticity assumptions, this optimisation problem can be explicitly solved with linear algebra by diagonalizing the variance-covariance matrix of the explanatory variables. 
In fine, we predict lane change according to the sign of the discriminant $D$.

\paragraph{Na\"ive Bayes classifier}
The na\"ive Bayes (NB) algorithm classifies the maneuver by maximizing the posterior conditional likelihood. % under independence and Gaussian assumptions. 
Using Bayes' theorem, the conditional posterior probability to observe a given maneuver is
$\mathbb P(Y|X) =\mathbb P(Y) \mathbb P(X|Y)/\mathbb P(X)$.
The probabilities $\mathbb P(Y)$ can be estimated using the empirical proportions
\begin{equation}
\tilde p_n=\frac1n\sum_{i=1}^n Y_i
\end{equation}
(see Table~\ref{tableStat}).
The conditional likelihood $\mathbb P(X|Y)$ is a complex function to optimize.
In the next, it is estimated assuming the measurements independent and normally distributed:
\begin{equation}
\tilde f(X,Y)=\prod_{k=1}^p \tilde f_k(X^k|Y),
\end{equation}
$\tilde f_k(\cdot|Y)$ being the empirical Gaussian density with two parameters $\mu_k$ and $\sigma_k$ conditionally to the maneuver.
The na\"ive Bayes classification is then the maneuver maximising the empirical posterior distribution
\begin{equation}
\tilde{Y}=\argmax_{Y,(\mu_k,\sigma_k)_k}
\tilde p_n(Y)\tilde f(X,Y).
\end{equation}

\paragraph{Predictive decision tree}
Decision trees (DT) are non-parametric supervised approaches. 
The CART algorithm is an iterative meta-heuristic consisting of successive partitions of the data maximizing the prediction accuracy  \cite{breiman1984classification}. 
This operation is repeated until the prediction improvement is marginal or leads to overfitting.
In contrast to discriminant analysis, na\"ive Bayes classifier, and, to a lesser extent, logistic model, the classification by decision tree is a non-linear approach. 
This also applies to the following two machine learning algorithms.

\paragraph{Support vector machine}
Support Vector Machine (SVM) is one of the most famous and robust supervised machine learning. 
The SVM algorithm maps training examples to points in space (the support vectors) maximizing the distance between two classes. 
In fine, the support vectors describe optimal hyper-planes partitioning the data by class. 
The classification, being linear by part, can be explicitly determined by linear algebra without requiring approximation meta-heuristics. 
This makes the approach computationally efficient, even for large data-set. 
SVMs can execute linear and partly-linear classification. 
The number of support vectors are hyper-parameters of the algorithm.

\paragraph{Artificial neural network}
Artificial neural networks (ANN) are machine learning methods based on nets of interconnected neurons. 
The neurons are activation functions, typically sigmoidal functions. 
Feed-forward ANNs are one of the simplest network architectures with no feedback mechanisms.
The neurons are organized in layers. 
The number of neurons and layers are hyper-parameters of the algorithms.
Too complex networks may lead to over-fitting phenomena.
A preliminary analysis shows that shallows networks allow obtaining precise and robust predictions. 
In the following, the ANN algorithms we compute consist of one layer with two neurons.

\subsection{Methodology}

We compare the prediction of the different models and algorithms using the mean square errors 
\begin{equation}
\begin{array}{lcl}
    \displaystyle\text{Total error}&=&\displaystyle\frac1N\sum_{i=1}^N(\hat Y_i-Y_i)^2\\[4mm]
    &=&\displaystyle\frac{N_\text{LK}}N\underbrace{\frac1{N_\text{LK}}\sum_{i,Y_i=0}^{N_\text{LK}}(\hat Y_i-Y_i)^2}_\text{\normalsize Error LK}
    +\frac{N_\text{LC}}N\underbrace{\frac1{N_\text{LC}}\sum_{i,Y_i=1}^{N_\text{LC}}(\hat Y_i-Y_i)^2}_\text{\normalsize Error LC}
    \end{array}
\end{equation}
with $Y_i$ and $\hat Y_i$ as the predicted and observed maneuvers (lane-changing or lane-keeping), respectively.
For a binary classification problem, the mean square error corresponds to the percent of prediction error. 
Three types of error are distinguished: The total error, the lane-keeping error (error LK, false negative), and the lane-changing error (error LC, false positive). 
Note that the total error is the average of lane-changing and lane-keeping errors weighted by the number of observations. 
We systematically cross-validate the results by sampling the data in training sets for the algorithm calibration, and testing sets for the prediction error computation. 
Eighty percent of the data are affected to the training. 
The twenty percent left is used for the testing. 
Cross-validation allows for controlling the overfitting of the machine learning algorithms.
The analysis is carried out over 1000 bootstrap training and testing sub-sampling.  
Bootstrap enables not only to obtain point-wise estimates for the errors but also to evaluate the precision of estimations \cite{kohavi1995study}. 
The bootstrap distributions determine whether a difference between two prediction errors is statistically significant.

\subsection{Benchmark analysis}

In this section, we present the lane-changing prediction results of the rule-based model MOBIL and the six data-based algorithms. 
The total error, the lane-changing error, and the lane-keeping error quantify the prediction accuracy. 
The testing and training errors are summarised in Table~\ref{tableallOV} for overtaking maneuvers, and in Table~\ref{tableallFD} for fold-down maneuvers. 

\begin{table}[!ht]\medskip\centering\small
\renewcommand{\arraystretch}{1.8}
\begin{tabular}{r|ccccccc}
& LR & LDA  & DT & SVM & NB & ANN&MOBIL\\
\hline
\small Tot.\ error& 6.3 \footnotesize(6.1)& 9.1 \footnotesize(8.9)& 7.3 \footnotesize(4.9)& 7.1 \footnotesize(6.3)& 9.2 \footnotesize(8.4)& {\bf 5.1} \footnotesize(4.1)& 9.9 \footnotesize(9.7)\\
\small Error LC& 31.0 \footnotesize(30.3)& 46.3 \footnotesize(45.4)& 31.5 \footnotesize(22.5)& 44.3 \footnotesize(39.7)& 56.1 \footnotesize(51.0)& {\bf 22.8} \footnotesize(19.1)& 54.5 \footnotesize(53.8)\\
\small Error LK&2.1 \footnotesize(2.0)& 2.8 \footnotesize(2.7)& 3.2 \footnotesize(1.9)& {\bf 0.8} \footnotesize(0.6)& 1.3 \footnotesize(1.1)& 2.2 \footnotesize(1.6)& 2.7 \footnotesize(2.7) 
\end{tabular}
\caption{Total, lane-changing and lane-keeping testing errors of the different prediction algorithms for the overtaking maneuver. The values between brackets are the training errors. Bold values indicate the lowest testing errors.}
\label{tableallOV}
\end{table}

\begin{table}[!ht]\centering\small
\renewcommand{\arraystretch}{1.8}
\begin{tabular}{r|ccccccc}
& LR & LDA  & DT & SVM & NB & ANN&MOBIL\\
\hline
\small Tot.\ error& 4.8 \footnotesize(4.7)& 5.1 \footnotesize(5.1)& 4.3 \footnotesize(3.5)& 4.3 \footnotesize(3.9)& 5.4 \footnotesize(5.3)& {\bf 4.2} \footnotesize(3.7)& 6.1 \footnotesize(6.0) \\      
\small Error LC&46.2 \footnotesize(45.9)& 41.6 \footnotesize(41.3)& 42.9 \footnotesize(36.9)& 50.9 \footnotesize(47.0)& 48.7 \footnotesize(48.0)& {\bf 33.3} \footnotesize(30.8)& 54.9 \footnotesize(54.0)\\
\small Error LK& 1.5 \footnotesize(1.4)& 2.2 \footnotesize(2.1)& 1.2 \footnotesize(0.8)& {\bf 0.5} \footnotesize(0.4)& 1.9 \footnotesize(1.8)& 1.9 \footnotesize(1.6)& 2.2 \footnotesize(2.1) 
\end{tabular}
\caption{Total, lane-changing and lane-keeping testing errors of the different prediction algorithms for the fold-down maneuver. The values between brackets are the training errors. Bold values indicate the lowest testing errors.}
\label{tableallFD}\medskip
\end{table}

\begin{figure}[!ht]
\centering
\input{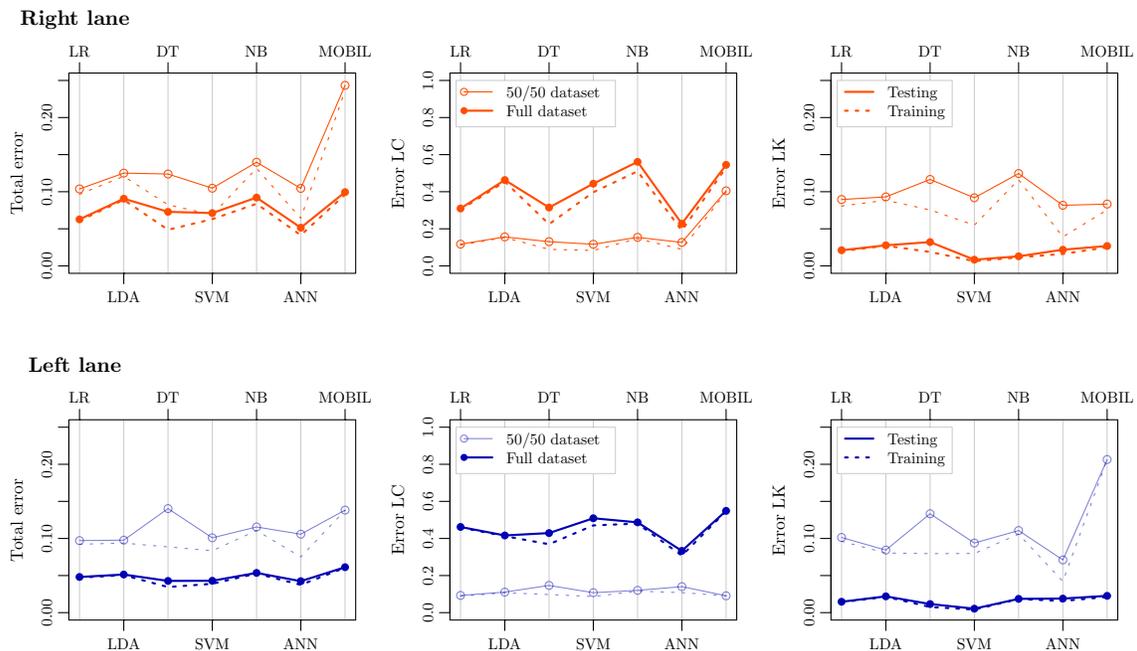}\\[-1mm]
\caption{Total, lane-changing, and lane-keeping testing and training errors for the six data-based algorithms and the MOBIL rule-based model. Top panel: Prediction on the right lane (lane-keeping or overtaking). Bottom panel: Prediction on the left lane (lane-keeping or fold-down). Two curves are provided: one for the full imbalanced data data-set and one for balanced data.}
\label{fig_Error}
\end{figure}

The average error is 7.7\% on the right lane (lane-keeping or overtaking maneuver) and 4,9\% on the left lane (lane-keeping or fold-down maneuver).
Generally speaking, the data-based algorithms, devoid of modeling bias, provide more accurate predictions. 
The averages of the total errors are 7.3\% and 4.7\% for the data-based algorithms on the right and left lanes. 
They are respectively for the rule-based model MOBIL 9.9\% and 6.1\%.
However, the predictions of lane-keeping maneuvers are more accurate than the predictions of lane-changing. 
The mean lane-keeping error is 2.2\% for overtaking and 1.4\% when folding down, while it is respectively 40.9\% and 45.5\% for the lane-changing maneuver. 
Indeed, the data are highly imbalanced (only approximately 9\% of the observations are lane-changing maneuvers). 
Using balanced data sets (by reducing the observations of lane-keeping maneuver) allows unifying the total, lane-changing, and lane-keeping errors to approximately 10\% except for the MOBIL model (see Fig.~\ref{fig_Error}).
Indeed, the MOBIL model is quite sensitive to the data balance. 
Maneuver predictions on the right lane especially weak with the balanced data set (error of 24.9\%).
The data-based algorithms are more robust, even if imbalanced effects still occur. 
The artificial neural network provides the most accurate predictions on both right (total error of 5.1\%, lane-changing 22.8\%, lane-keeping 2.2\%) and left lanes (total error of 4.2\%, lane-changing 33.3\%, lane-keeping 1.9\%). 
The prediction accuracy for the lane-changing maneuver is relatively modest. 
However, we remain that it is only based on single observations of relative speeds and distances to the neighbors two seconds before the lane-changing occurs.

\begin{figure}[!ht]
\centering\vspace{-10mm}
\input{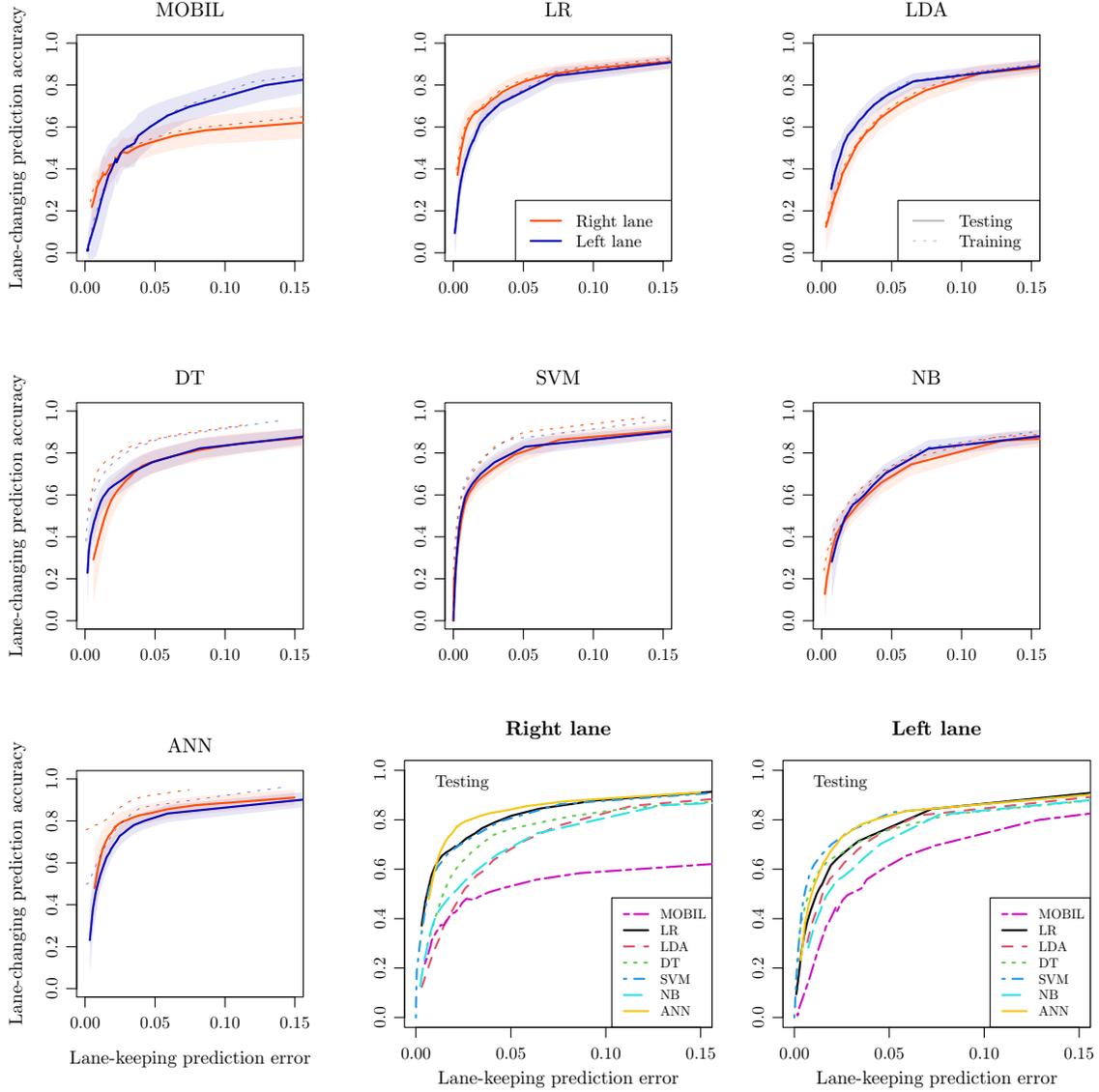}\\[2mm]
\caption{Receiver operating characteristic (ROC) curves for the data-based algorithms and the MOBIL rule-based model obtained by sub-sampling the data during the training phase. The data-based algorithms systematically outperform the predictions of the rule-based model. The best maneuver prediction results from the neural network for both right and left lanes.}
\label{figROCplot}
\end{figure}

The initial data-set is highly imbalanced with few observations of lane-changing maneuvers.
The receiver operating characteristic (ROC) curves present the lane-changing prediction accuracy (true positive) according to the lane-keeping prediction error (false positive) by sub-sampling the data balance during the training phase. 
The ROC curves allow obtaining an overview of the prediction ability of the approaches independent of the data bias. 
In Fig.~\ref{figROCplot}, the orange ROC curves present maneuver prediction in the right lane, while the blue curves correspond to maneuver prediction in the left lane.
The continuous curves correspond to the testing errors while the dotted curves are the training results.
The data-based algorithms systematically over-perform the predictions of the MOBIL rule-based model. 
ANNs, SVMs, and, to a lesser extent, logistic regressions show higher prediction accuracy.
Note that decision trees, SVMs, and ANNs are especially subject to overfitting phenomena with higher accuracy in training phases.

\subsection{Ensemble learning algorithms}

Ensemble learning techniques use a finite ensemble of algorithms to improve prediction performance. 
In this section, we aim to combine the six data-based algorithms previously individually used to predict the maneuvers. 
We compare ten ensemble learning techniques relying on bootstrap aggregating (bagging) and stacking meta-heuristics.
The bagging meta-heuristics consist of 
\begin{itemize}
\item The max and min consensus rules: The prediction is lane-changing as soon as \textit{at least one} of the six single algorithms predicts a lane change (max), while it occurs when \textit{all} the algorithms predict a lane change (min). 
\item The mean algorithm: A majority rule over the six individual algorithm predictions (the classification is random in case of equality).
\item The weighted mean algorithm (mean$^\star$): The average of the six individual algorithm predictions weighted by the prediction accuracy.
\end{itemize}
The stacking meta-heuristics is a tow-step classification method taking as inputs the explanatory variables $X$ as well as the predictions of the six individual data-based algorithms. 
The meta-heuristic used for the final classification is one of the  data-based algorithms, namely logistic regression (LR), linear discriminant (LDA), decision tree (DT), na\"ive Bayes classifier (NB), support-vector-machine (SVM), or shallow artificial neural network (ANN).

\begin{figure}[!b]
\centering
\input{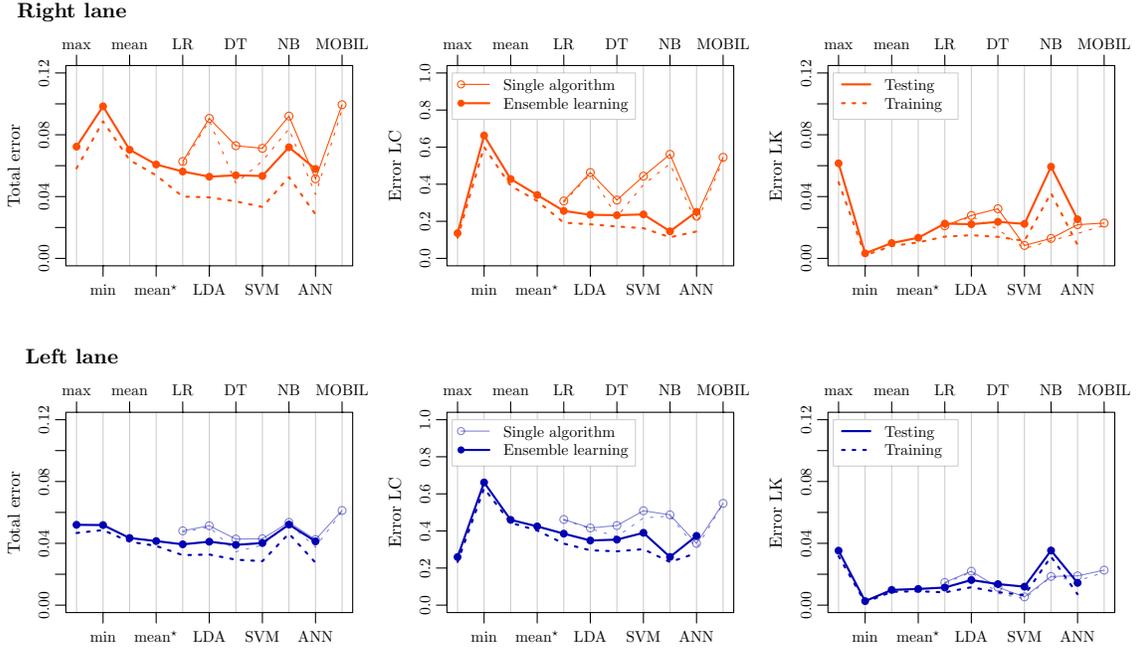}\\[-1mm]
\caption{Total, lane-changing, and lane-keeping errors for the six single data-based algorithms and the ten bagging and stacking ensemble learning meta-heuristics. Top panel: Prediction on the right lane (lane-keeping or overtaking). Bottom panel: Prediction on the left lane (lane-keeping or fold-down).}
\label{fig_Error_Ens}
\end{figure}

\begin{figure}[!ht]
\centering\small\vspace{0mm}
\input{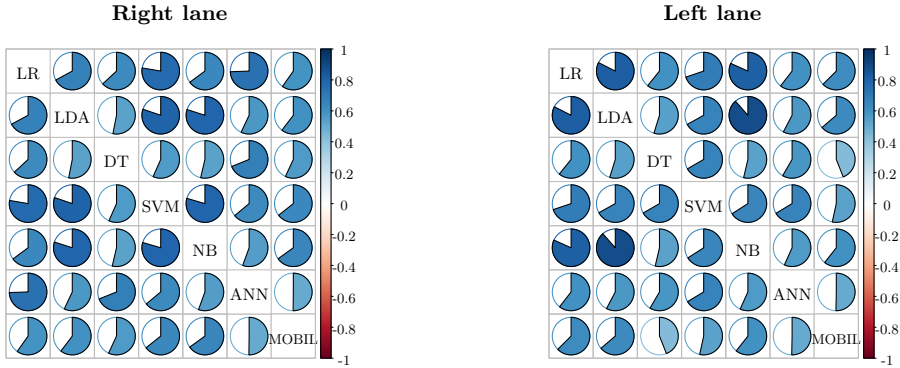}\vspace{-4mm}
\caption{Table of correlation of the six machine learning algorithm and the MOBIL model predictions on right (left panel) and left lane (right panel). The predictions are in general highly correlated.}\label{figCorrML}
\end{figure}

We compute the training and testing errors using the ten ensemble learning techniques over the full (imbalanced) data-set.
The mean errors are compared to the single data-based algorithm errors in Fig.~\ref{fig_Error_Ens}. 
Beyond the bagging approaches, the weighted average mean$^\star$ provides the most accurate prediction. 
The consensus rules allow minimizing the lane-changing error at the expense of the lane-keeping error (max algorithm), and conversely (min algorithm). 
The resulting total error is increased (especially for the min algorithm). 
Except for the na\"ive Bayes classifier, the prediction errors with the six stacking meta-heuristics are relatively uniform on both the right and left lanes.
The best total errors are approximately 5\% in the right lane and 4\% in the left lane. 
These estimates are close to those of the weighted average bagging technique or to the best single algorithm (the artificial neural network ANN, see Tables~\ref{tableallOV} and \ref{tableallFD}).
The stacking meta-algorithms allow uniforming the lane-changing and lane-keeping errors as well. 
Furthermore, the algorithm diversity compensates for the data imbalance by significantly reducing the lane-changing error. 
This does not lead, however, to a significant lane-keeping error increase.
We might have expected more accurate predictions by using ensemble learning meta-heuristics that combine up to six algorithms. 
However, it turns out that the machine learning lane-changing prediction errors are highly correlated (see Fig.~\ref{figCorrML}). 
The mean prediction error correlation is 0.64 on the right lane and 0.63 on the left lane. 
The prediction error correlation strongly limits the performance of the ensemble learning redundancy. 
Such phenomenon has recently been empirically pointed out for multi-sensor perception of autonomous vehicles \cite{gottschalk2022does}.

\subsection{Early predictions} 

\begin{figure}[!ht]
\centering\vspace{-7mm}
\input{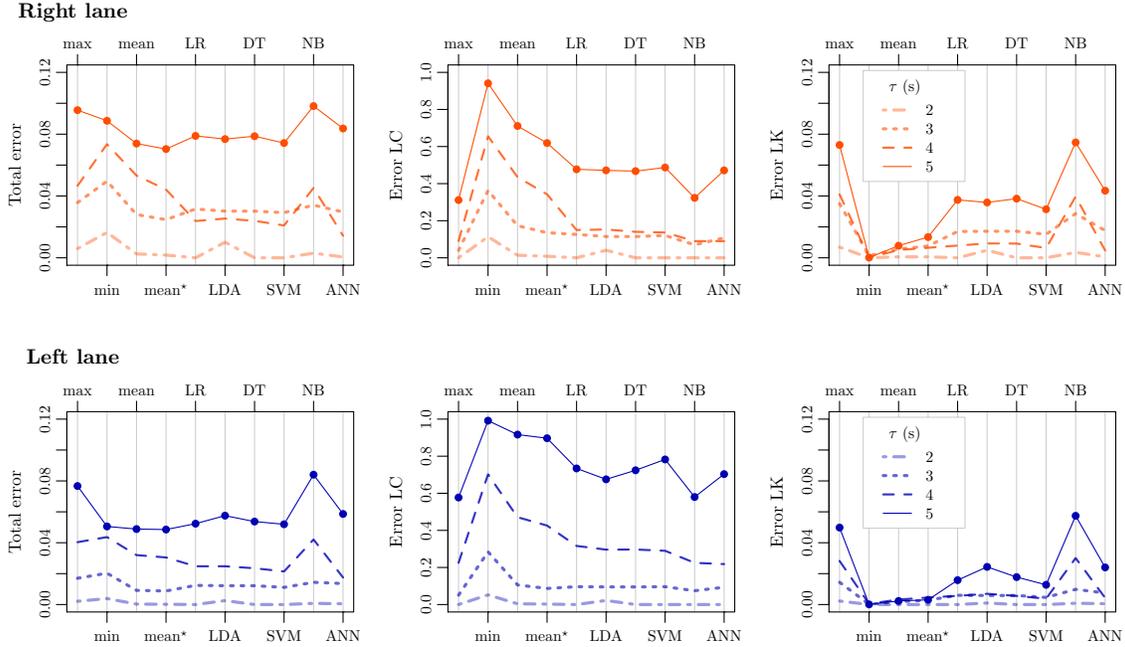}\\[-1mm]
\caption{Total, lane-changing, and lane-keeping errors with the bagging and stacking ensemble learning techniques for prediction horizon time ranging from two to five seconds before the lane-changing occurs. Top panel: Prediction on the right lane (lane-keeping or overtaking). Bottom panel: Prediction on the left lane (lane-keeping or fold-down). The ANN-stacking meta-heuristic provides predictions four seconds in advance with total errors of 2\% (overtaking, fold-down and lane-keeping errors are 12.2\%, 20.5\%, and 0.5\%, respectively).}
\label{fig_Error_Ens_diffTau}
\end{figure}

Ensemble learning techniques are particularly efficient for large multivariate data sets. 
In this section, we aim to predict early lane-changing maneuvers up to five seconds in advance using the twenty-four measurements with the surrounding vehicles (see Fig.~\ref{figPredicators} and Table~\ref{tablePredicators}). 
Figure~\ref{fig_Error_Ens_diffTau} presents the prediction errors of the ten bagging and stacking ensemble learning meta-heuristics introduced above. 
We perform predictions for horizon time from two to five seconds before the vehicle crosses the highway central line. 
The predictions with the twenty-four variables over two seconds are trivial and obtained with almost no error for all tested meta-heuristics. 
This is due to the usage of lateral velocity and acceleration, for which high values almost certainly indicate a lane change. 
The variation of the lateral movement is observed up to two seconds before the vehicle crosses the central line. 
Surprisingly, the stacking methods may even allow more accurate predictions over four than three seconds in the right lane. 
The accuracy of the neural network meta-heuristic for prediction over four seconds is high.
Indeed, the total errors are approximately 2\% on both left and right lanes (overtaking, fold-down and lane-keeping errors are 12.2\%, 20.5\%, and 0.5\%, respectively).

\section{Summary and conclusion}
\label{section5}

In this article, we empirically analyse and predict lane-changing maneuvers on two-lane highways using HighD trajectory data-set \cite{highd}. 
We separately analyze lane-changing maneuvers in the right and left lanes (overtaking and fold-down maneuvers). 
The empirical analysis shows that the situations intending to overtake or fold down on European highways are different and even, somehow, opposed. 
Overtaking maneuvers mainly result from mechanisms based on speed difference and distance to the current predecessor and adjacent follower. 
Fold-down maneuvers are more complex processes involving additional surrounding vehicles and variables.

The prediction results demonstrate that machine and ensemble learning techniques can accurately forecast highway lane-changing maneuvers from instantaneous local time-space measurements. 
Specifically, we observe that the data-based approaches systematically outperform the rule-based model MOBIL, even when the inputs (i.e., spacing and speed difference with four surrounding vehicles) are identical. 
Among the machine learning algorithms, we observe that artificial neural networks, support vector machines, and, to a lesser extent, logistic regressions provide the most accurate predictions. 
Ensemble learning meta-heuristics are particularly powerful for early predictions when the number of variables measured is large. 
Stacking meta-heuristics allow accurate predictions up to four seconds before the lane-changing occurs, with total prediction errors of approximately 2\% on both left and right lanes (overtaking, fold-down and lane-keeping errors being 12.2\%, 20.5\%, and 0.5\%, respectively). 
The algorithm prediction errors are highly correlated, limiting the ensemble learning performances.
 
Even if the predictions may have good accuracy, it remains to analyze the situations for which the algorithms fail. 
This is especially relevant for wrong predictions of overtaking or fold-down maneuvers (i.e., lane-keeping error). 
Indeed, a wrong lane-changing prediction may generate an accident in case of the presence of a vehicle in the adjacent lane. 
For safety reasons, data-driven algorithms must be coupled with minimal collision-free criteria based on kinematic principles.
The applications rely on extensions of adaptive cruise control systems, including lane change maneuver predictions.
However, the setting up, training, and reproducibility of the machine learning predictions remains to be tested and evaluated. 
This will be the topics of future works.

\section*{Data availability}
The data that support the findings of this study are available from the authors upon request.

\section*{Declaration of interests}
The authors declare that they have no known competing financial interests or personal relationships that could have appeared to influence the work reported in this paper.

\section*{Acknowledgement}
The authors acknowledge financial support through the research consortium Bergisch.Smart\_- Mobility funded by the Ministry for Economy, Innovation, Digitalization and Energy (MWIDE) of the state North Rhine Westphalia under the grant no.\ DMR-3-2.

\end{document}